\documentclass[Alon2,ChapterTOCs,singlecolor,11pt]{Alon}

\usepackage{fixltx2e,fix-cm}
\usepackage[english]{babel}
\usepackage{epigrafica}
\usepackage[LGR,OT1]{fontenc}
\usepackage{lmodern}
\usepackage{amssymb}
\usepackage{amsmath}
\usepackage{graphicx}
\usepackage{subfigure}
\usepackage{makeidx}
\usepackage{multicol}
\usepackage{url}
\usepackage{multirow}
\usepackage{tabularx}
\usepackage{changepage}
\usepackage{version}
\usepackage[super]{nth}

\title{Control Strategies for Autonomous Vehicles}

\begin{document}

\chapter[Control Strategies for Autonomous Vehicles]{\centering \Huge CONTROL STRATEGIES FOR AUTONOMOUS VEHICLES}

\pagebreak

\chapterinitial{T}{HIS CHAPTER} shall focus on the self-driving technology from a control perspective and investigate the control strategies used in autonomous vehicles and advanced driver-assistance systems (ADAS) from both theoretical and practical viewpoints.

First, we shall introduce the self-driving technology as a whole, including perception, planning and control techniques required for accomplishing the challenging task of autonomous driving. We shall then dwell upon each of these operations to explain their role in the autonomous system architecture, with a prime focus on control strategies.

The core portion of this chapter shall commence with detailed mathematical modeling of autonomous vehicles followed by a comprehensive discussion on control strategies. The chapter shall cover longitudinal as well as lateral control strategies for autonomous vehicles with coupled and de-coupled control schemes. We shall as well discuss some of the machine learning techniques applied to autonomous vehicle control task.

Finally, there shall be a brief summary of some of the research works that our team has carried out at the Autonomous Systems Lab (SRMIST) and the chapter shall conclude with some thoughtful closing remarks.

\section{Introduction} \label{Introduction}

Autonomous vehicles, or self-driving cars as they are publicly referred to, have been the dream of mankind for decades. This is a rather complex problem statement to address and it requires inter-disciplinary expertise, especially considering the safety, comfort and convenience of the passengers along with the variable environmental factors and highly stochastic fellow agents such as other vehicles, pedestrians, etc. The complete realization of this technology shall, therefore, mark a significant step in the field of engineering.

\subsection{The Autonomous Driving Technology} \label{The Autonomous Driving Technology}

Autonomous vehicles perceive the environment using a comprehensive sensor suite and process the raw data from the sensors in order to make informed decisions. The vehicle then plans the trajectory and executes controlled maneuver in order to track the trajectory autonomously. This process is elucidated below.

\begin{figure}[htb]
	
	\centering
	\includegraphics[width=\textwidth]{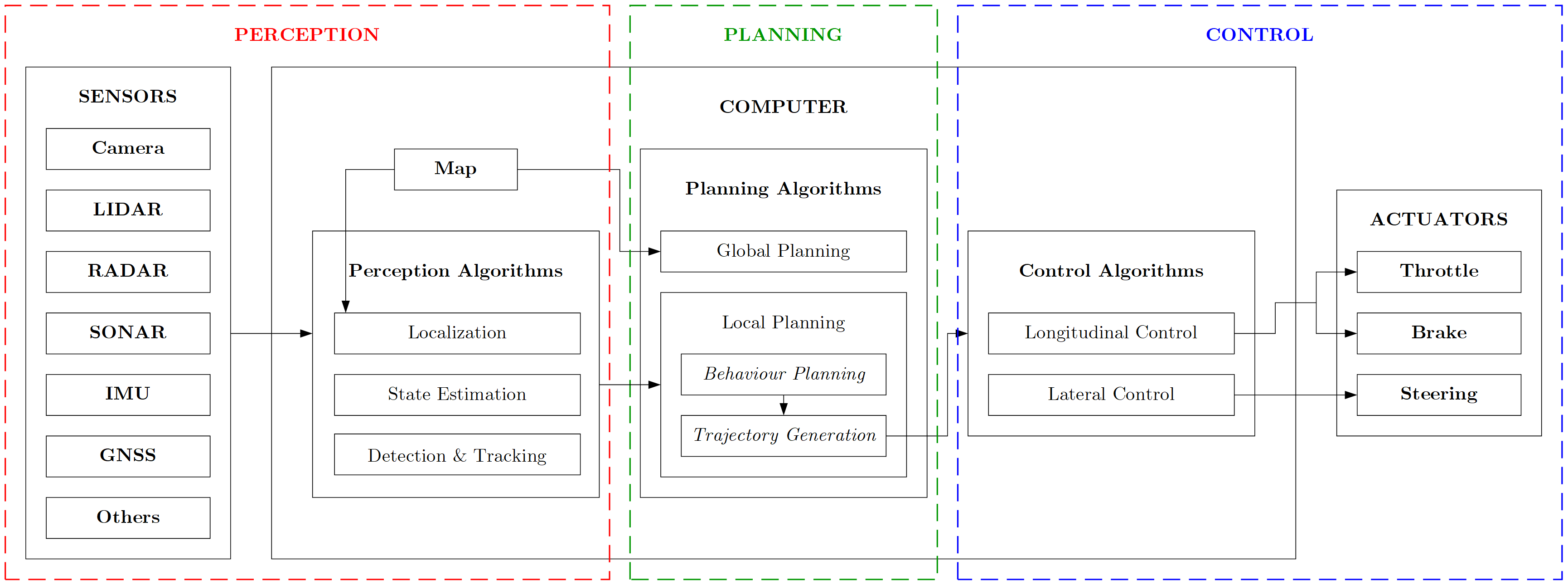}
	\caption{System Architecture of an Autonomous Vehicle}
	\label{fig:System Architecture of an Autonomous Vehicle}
	
\end{figure}

The system architecture of an autonomous vehicle is fundamentally laid out as a stack that constitutes of 3 sub-systems as shown in figure \ref{fig:System Architecture of an Autonomous Vehicle}.

\begin{enumerate}
	
	\item \textbf{Perception:} Autonomous vehicles employ various types of sensors such as cameras, LIDARs, RADARs, SONARs, IMU, GNSS, etc. in order to perceive the environment (object and event detection and tracking) and monitor their own physical parameters (localization and state estimation) for making informed decisions. Suitable sensor fusion and filtering algorithms including simpler ones such as complementary filter, moving average filter, etc. and advanced ones such as Kalman filter (and its variants), particle filter, etc. are adopted in order to reduce measurement uncertainty due to noisy sensor data. In a nutshell, this sub-system is responsible for providing both intrinsic and extrinsic knowledge to the ego vehicle using various sensing modalities.
	
	\item \textbf{Planning:} Using the data obtained from the perception sub-system, the ego vehicle performs behavior planning wherein the most optimal behavior to be adopted by the ego vehicle needs to be decided by predicting states of the ego vehicle as well as other dynamic objects in the environment into the future by certain prediction horizon. Based on the planned behavior, the motion planning module generates an optimal trajectory, considering the global plan, passenger safety, comfort as well as hard and soft motion constraints. This entire process is termed as \textit{motion planning}. Note that there is a subtle difference between path planning and motion planning in that the prior is responsible only for planning the reference ``path" to a goal location (ultimate or intermediate goal) whereas the later is responsible for planning the reference ``trajectory" to a goal location (i.e. considers not just the pose, but also its higher derivatives such as velocity, acceleration and jerk, thereby assigning a temporal component to the reference path).
	
	\item \textbf{Control:} Finally, the control sub-system is responsible for accurately tracking the trajectory provided by the planning sub-system. It does so by appropriately adjusting the final control elements (throttle, brake and steering) of the vehicle.
	
\end{enumerate}

\subsection{Significance of Control System} \label{Significance of Control System}

A control system is responsible for regulating or maintaining the process conditions of a plant at their respective desired values by manipulating certain process variable(s) to adjust the variable(s) of interest, which is/are generally the output variable(s).

If we are to look from the perspective of autonomous vehicles, the control system is dedicated to generate appropriate commands for throttle, brake and steering (input variables) so that the vehicle (plant) tracks a prescribed trajectory by executing a controlled motion (where motion parameters such as position, orientation, velocity, acceleration, jerk, etc. are the output variables). It is to be noted that the input and output variables (a.k.a. manipulated/process variables and controlled variables, respectively) are designated so with respect to the plant and not the controller, which is a common source of confusion.

Control system plays a very crucial role in the entire architecture of an autonomous vehicle and being the last member of the pipeline, it is responsible for actually ``driving" the vehicle. It is this sub-system, which ultimately decides how the ego vehicle will behave and interact with the environment.

Although the control sub-system cannot function independently without the perception and planning sub-systems, it is also a valid argument that the perception and planning sub-systems are rendered useless if the controller is not able to track the prescribed trajectory accurately.

\subsection{Control System Architecture for Autonomous Vehicles} \label{Control System Architecture for Autonomous Vehicles}

The entire control system of an autonomous vehicle is fundamentally broken down into the following two components:

\begin{itemize}
	
	\item \textbf{Longitudinal Control:} This component controls the longitudinal motion of the ego vehicle, considering its longitudinal dynamics. The controlled variables in this case are throttle and brake inputs to the ego vehicle, which govern its motion (velocity, acceleration, jerk and higher derivatives) in the longitudinal direction.
	
	\item \textbf{Lateral Control:} This component controls the lateral motion of the ego vehicle, considering its lateral dynamics. The controlled variable in this case is the steering input to the ego vehicle, which governs its steering angle and heading. Note that steering angle and heading are two different terminologies. While steering angle describes the orientation of the steerable wheels and hence the direction of motion of the ego vehicle, heading is concerned with the orientation of the ego vehicle.
	
\end{itemize}

\section{Mathematical Modeling} \label{Mathetical Modeling}

Mathematical modeling of a system refers to the notion of describing the response of a system to the control inputs while accounting the state of the system using mathematical equations. The following analogy of an autonomous vehicle better explains this notion.

When control inputs are applied to an autonomous vehicle, it moves in a very specific way depending upon the control inputs. For example, throttle increases the acceleration, brake reduces it, while steering alters the heading of the vehicle by certain amount. A mathematical model of such an autonomous vehicle will represent the exact amount of linear and/or angular displacement, velocity, acceleration, etc. of the vehicle depending upon the amount of applied throttle, brake and/or steering input(s).

In order to actually develop a mathematical model of autonomous vehicle (or any system for that matter), there are two methods widely used in industry and academia.

\begin{enumerate}
	
	\item \textbf{First Principles Modeling:} This approach is concerned with applying the fundamental principles and constituent laws to derive the system models. It is a theoretical way of dealing with mathematical modeling, which does not necessarily require access to the actual system and is mostly adopted for deducing generalized mathematical models of the concerned system. We will be using this approach in the upcoming section to formulate the kinematic and dynamic models of a front wheel steered non-holonomic (autonomous) vehicle.
	
	\item \textbf{Modeling by System Identification:} This approach is concerned with applying known inputs to the system, recording it's responses to those inputs and statistically analyzing the input-output relations to deduce the system models. This approach is a practical way of dealing with mathematical modeling, which requires access to the actual system and is mostly adopted for modeling complex systems, especially where realistic system parameters are to be captured. It is to be noted that this approach is often helpful to estimate system parameters even though the models are derived using first principles approach.
	
\end{enumerate}

System models can vary from very simplistic linear models to highly detailed and complex, non-linear models. The complexity of model being adopted depends on the problem at hand.

There is always a trade-off between model accuracy and the computational complexity that comes with it. Owing to their accuracy, complex motion models may seem attractive at the first glance; however, they may consume a considerable amount of time for computation, making them not so ``real-time" executable. When it comes to safety-critical systems like autonomous vehicles, latency is to be considered seriously as failing to perform actions in real-time may lead to catastrophic consequences. Thus in practice, for systems like autonomous vehicles, often approximate motion models are used to represent the system dynamics. Again, the level of approximation depends on factors like driving speed, computational capacity, quality of sensors and actuators, etc.

It is to be noted here that the models describing vehicle motion are not only useful in control system design, but are also a very important tool for predicting future states of ego vehicle or other objects in the scene (with associated uncertainty), which is extremely useful in the perception and planning phases of autonomous vehicles.

We make the following assumptions and consider the following motion constraints for modeling the vehicle.

\textbf{\\Assumptions:}
\begin{enumerate}
	\item The road surface is perfectly planar, any elevations or depressions are disregarded. This is known as the \textit{planar assumption}.
	\item Front and rear wheels are connected by a rigid link of fixed length.
	\item Front wheels are steerable and act together, and can be effectively represented as a single wheel.
	\item Rear wheels act together and can be effectively represented as a single wheel.
	\item The vehicle is actually controllable like a bicycle.
\end{enumerate}

\textbf{\\Motion Constraints:}
\begin{enumerate}
	\item \textbf{Pure Rolling Constraint:} This constraint implies the fact that each wheel follows a pure rolling motion w.r.t. ground; there is no slipping or skidding of the wheels.
	\item \textbf{Non-Holonomic Constraint:} This constraint implies that the vehicle can move only along the direction of heading and cannot arbitrarily slide along the lateral direction.
\end{enumerate}

\subsection{Kinematic Model} \label{Kinematic Model}

Kinematics is the study of motion of a system disregarding the forces and torques that govern it. Kinematic models can be employed in situations wherein kinematic relations are able to sufficiently approximate the actual system dynamics. It is important to note, however, that this approximation holds true only for systems that perform non-aggressive maneuvers at lower speeds. To quote an example, kinematic models can nearly-accurately represent a vehicle driving slowly and making smooth turns. However, if we consider something like a racing car, it is very likely that the kinematic model would fail to capture the actual system dynamics.

In this section, we present one of the most widely used kinematic model for autonomous vehicles, the \textit{kinematic bicycle model}. This model performs well at capturing the actual vehicle dynamics under nominal driving conditions. In practice, this model tends to strike a good balance between simplicity and accuracy and is therefore widely adopted. That being said, one can always develop more detailed and complex models depending upon the requirement.

The idea is to define the vehicle state and see how it evolves over time based on the previous state and current control inputs given to the vehicle.

\begin{figure}[htb]
	
	\centering
	\includegraphics[width=\textwidth]{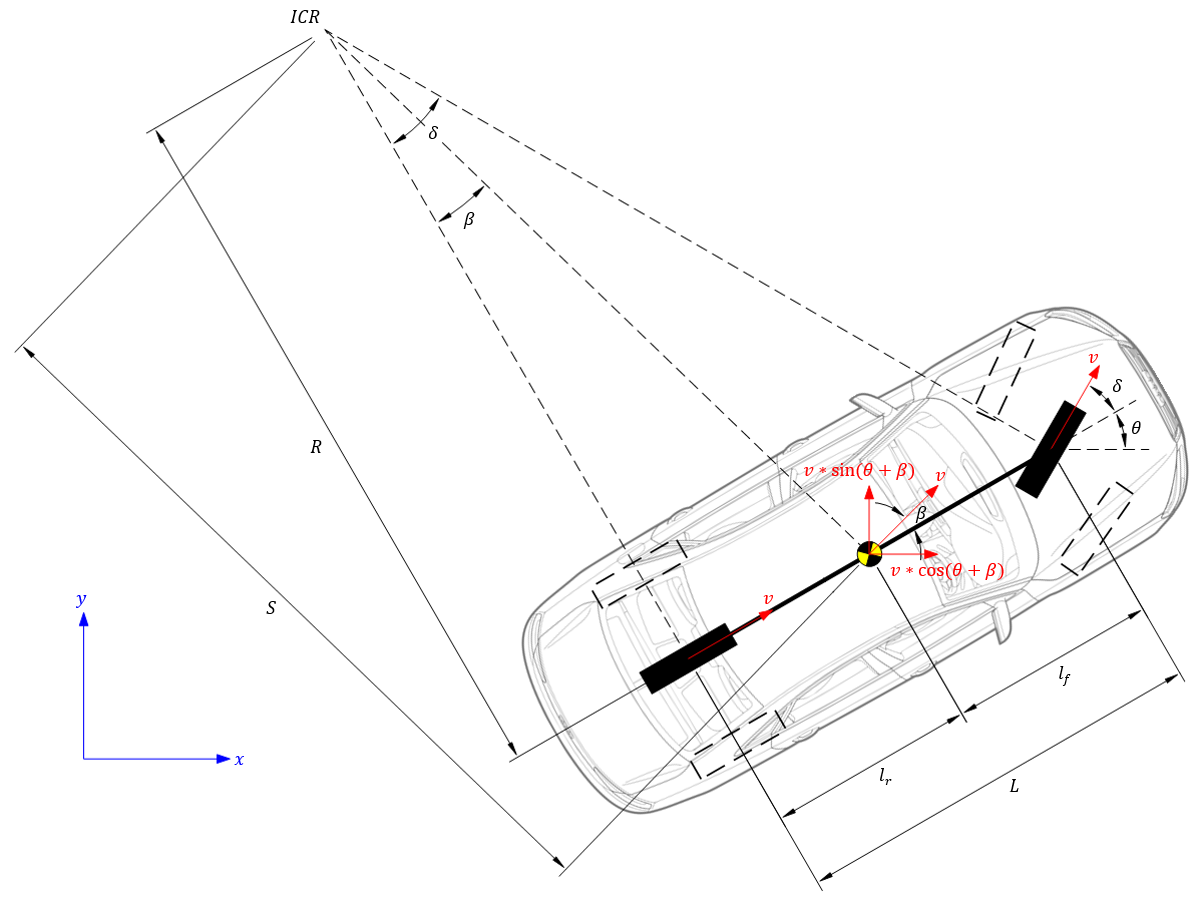}
	\caption{Vehicle Kinematics}
	\label{fig:Vehicle Kinematics}
	
\end{figure}

Let the vehicle state constitute $x$ and $y$ components of position, heading angle or orientation $\theta$ and velocity (in the direction of heading) $v$. Summarizing, ego vehicle state vector $q$ is defined as follows.

\begin{equation}\label{Equation 3.1}
q=\left[x,y,\theta,v\right]^T
\end{equation}

For control inputs, we need to consider both longitudinal (throttle and brake) and lateral (steering) commands. The brake and throttle commands contribute to longitudinal accelerations in range of $\left[-a'_{max},a_{max}\right]$ where negative values represent deceleration due to braking and positive values represent acceleration due to throttle (forward or reverse depending upon the transmission state). Note that the limits $a'_{max}$ and $a_{max}$ are intentionally denoted so as to distinctly illustrate the difference between the physical limits of acceleration due to throttle and deceleration due to braking. The steering command alters the steering angle $\delta$ of the vehicle, where $\delta \in \left[-\delta_{max},\delta_{max}\right]$ such that negative steering angles dictate left turns and positive steering angles otherwise. Note that generally, the control inputs are clamped in the range of $\left[-1,1\right]$ based on the actuators for proper scaling of control commands in terms of actuation limits. Summarizing, ego vehicle control vector $u$ is defined as follows.

\begin{equation}\label{Equation 3.2}
u=\left[a,\delta\right]^T
\end{equation}

We will now derive the kinematic model of the ego vehicle. As shown in figure \ref{fig:Vehicle Kinematics}, the local reference frame is located at the center of gravity of the ego vehicle.

Using the distance between rear wheel axle and vehicle's center of gravity, we can compute the slip angle $\beta$ as follows.

\begin{equation*}
tan\left(\beta\right)=\frac{l_r}{S}=\frac{l_r}{\left(\frac{L}{tan\left(\delta\right)}\right)}=\frac{l_r}{L}*tan\left(\delta\right)
\end{equation*}

\begin{equation}\label{Equation 3.3}
\therefore\beta=tan^{-1}\left(\frac{l_r}{L}*tan\left(\delta\right)\right)
\end{equation}

\begin{equation*}
\text{Ideally, $l_r=L/2$} \Rightarrow\beta=tan^{-1}\left(\frac{tan\left(\delta\right)}{2}\right)
\end{equation*}

Resolving the velocity vector $v$ into $x$ and $y$ components using the laws of trigonometry we get,

\begin{equation}\label{Equation 3.4}
\dot{x}=v*cos\left(\theta+\beta\right)
\end{equation}

\begin{equation}\label{Equation 3.5}
\dot{y}=v*sin\left(\theta+\beta\right)
\end{equation}

In order to compute $\dot{\theta}$, we first need to calculate $S$ using the following relation.

\begin{equation}\label{Equation 3.6}
S=\frac{L}{tan\left(\delta\right)}
\end{equation}

Using $S$ obtained from equation \ref{Equation 3.6}, we can compute $R$ as given below.

\begin{equation}\label{Equation 3.7}
R=\frac{S}{cos\left(\beta\right)}=\frac{L}{\left(tan\left(\delta\right)*cos\left(\beta\right)\right)}
\end{equation}

Using $R$ obtained from equation \ref{Equation 3.7}, we can deduce $\dot{\theta}$ as follows.

\begin{equation}\label{Equation 3.8}
\dot{\theta}=\frac{v}{R}=\frac{v*tan\left(\delta\right)*cos\left(\beta\right)}{L}
\end{equation}

Finally, we can compute $\dot{v}$ using the rudimentary differential relation.
\begin{equation}\label{Equation 3.9}
\dot{v}=a
\end{equation}

Using equations \ref{Equation 3.4}, \ref{Equation 3.5}, \ref{Equation 3.8} and \ref{Equation 3.9}, we can formulate the continuous-time kinematic model of autonomous vehicle.

\begin{equation}\label{Equation 3.10}
\dot{q}=\begin{bmatrix}\dot{x}\\ \dot{y}\\ \dot{\theta}\\\dot{v}\end{bmatrix}=\begin{bmatrix}v*cos\left(\theta+\beta\right)\\ v*sin\left(\theta+\beta\right)\\ \frac{v*tan\left(\delta\right)*cos\left(\beta\right)}{L}\\ a\end{bmatrix}
\end{equation}

Based on the formulation in equation \ref{Equation 3.10}, we can formulate the discrete-time model of autonomous vehicle.

\begin{equation}\label{Equation 3.11}
\left\{\begin{matrix}x_{t+1}=x_t+\dot{x_t}*\Delta t\\ y_{t+1}=y_t+\dot{y_t}*\Delta t\\ \theta_{t+1}=\theta_t+\dot{\theta_t}*\Delta t\\ v_{t+1}=v_t+\dot{v_t}*\Delta t\end{matrix}\right.
\end{equation}

Note that the equation \ref{Equation 3.11} is known as the \textit{state transition equation} (generally represented as $q_{t+1}=q_t+\dot{q_t}*\Delta t$) where $t$ in the subscript denotes current time instant and $t+1$ in the subscript denotes next time instant.

\subsection{Dynamic Model} \label{Dynamic Model}

Dynamics is the study of motion of a system with regard to the forces and torques that govern it. In other words, dynamic models are motion models of a system that closely resemble the actual system dynamics. Such models tend to be more complex and inefficient to solve in real-time (depends on computational hardware) but are much of a requirement for high-performance scenarios such as racing, where driving at high speeds and with aggressive maneuvers is common.

There are two popular methodologies for formulating dynamic model of systems viz. Newton-Euler method and Lagrange method. In Newton-Euler approach, one has to consider all the forces and torques acting on the system, whereas in the Lagrange approach, the forces and torques are represented in terms of potential and kinetic energies of the system. It is to be noted that both approaches are equally correct and result in equivalent formulation of dynamic models. In this section, we will present longitudinal and lateral dynamic models of an autonomous vehicle using Newton-Euler approach.

\subsubsection{Longitudinal Vehicle Dynamics}

\begin{figure}[htb]
	
	\centering
	\includegraphics[width=\textwidth]{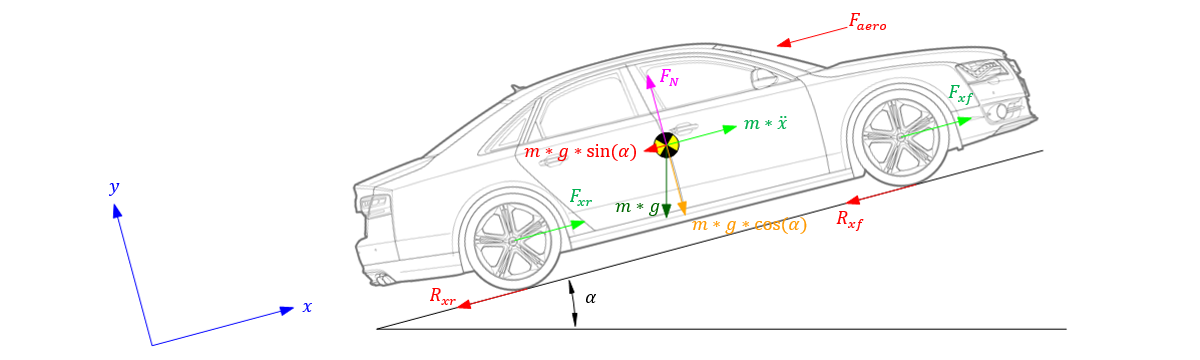}
	\caption{Longitudinal Vehicle Dynamics}
	\label{fig:Longitudinal Vehicle Dynamics}
	
\end{figure}

The free-body diagram of an autonomous vehicle along the longitudinal direction (denoted as $x$) is depicted in figure \ref{fig:Longitudinal Vehicle Dynamics}. The longitudinal forces considered include vehicle inertial term $m*\ddot{x}$, front and rear tire forces $F_{xf}$ and $F_{xr}$, aerodynamic resistance $F_{aero}$, front and rear rolling resistance $R_{xf}$ and $R_{xr}$, and $x$ component of the gravitational force $m*g*sin\left(\alpha\right)$ (since $y$ component of the gravitational force $m*g*cos\left(\alpha\right)$ and normal force $F_N$ cancel each other). Note that the tire forces assist the vehicle to move forward whereas all other forces resist the forward vehicle motion.

Thus, applying Newton's second law of motion to the free-body diagram we get,

\begin{equation*}
m*\ddot{x}=F_{xf}+F_{xr}-F_{aero}-R_{xf}-R_{xr}-m*g*sin\left(\alpha\right)
\end{equation*}

The above dynamic equation can be simplified as follows. Let front and rear tire forces collectively represent traction force $F_x$ and the front and rear rolling resistance collectively represent net rolling resistance $R_x$. Also, making a small angle approximation for $\alpha$, we can say that $sin\left(\alpha\right)\approx\alpha$. Therefore we have,

\begin{equation*}
m*\ddot{x}=F_x-F_{aero}-R_x-m*g*\alpha
\end{equation*}

\begin{equation}\label{Equation 3.12}
\therefore \ddot{x}=\frac{F_x}{m}-\frac{F_{aero}}{m}-\frac{R_x}{m}-g*\alpha
\end{equation}

The individual forces can be modeled as follows.

\begin{itemize}
	
	\item \textbf{Traction force} $F_x$ depends upon vehicle mass $m$, wheel radius $r_{wheel}$ and angular acceleration of the wheel $\ddot{\theta}_{wheel}$. Since $F=m*a$ and $a=r*\ddot{\theta}$, we have,
	\begin{equation}\label{Equation 3.13}
	F_x=m*r_{wheel}*\ddot{\theta}_{wheel}
	\end{equation}
	
	\item \textbf{Aerodynamic resistance} $F_{aero}$ depends upon air density $\rho$, frontal surface area of the vehicle $A$ and velocity of the vehicle $v$. Using proportionality constant $C_\alpha$ we have,
	\begin{equation}\label{Equation 3.14}
	F_{aero}=\frac{1}{2}*C_\alpha*\rho*A*v^2\approx C_\alpha*v^2
	\end{equation}
	
	\item \textbf{Rolling resistance} $R_x$ depends upon tire normal force $N$, tire pressure $P$ and velocity of the vehicle $v$. Note that tire pressure is a function of vehicle velocity.
	\begin{equation*}
	R_x=N*P\left(v\right)
	\end{equation*}
	Where, $P\left(v\right)=\hat{C}_{r,0}+\hat{C}_{r,1}*\left|v\right|+\hat{C}_{r,2}*v^2$
	\begin{equation}\label{Equation 3.15}
	\therefore R_x=N*\left(\hat{C}_{r,0}+\hat{C}_{r,1}*\left|v\right|+\hat{C}_{r,2}*v^2\right)\approx \hat{C}_{r,1}*\left|v\right|
	\end{equation}
	
\end{itemize}

Substituting equations \ref{Equation 3.13}, \ref{Equation 3.14} and \ref{Equation 3.15} in equation \ref{Equation 3.12} we get,

\begin{equation}\label{Equation 3.16}
\ddot{x}=r_{wheel}*\ddot{\theta}_{wheel}-\frac{C_\alpha*v^2}{m}-\frac{\hat{C}_{r,1}*\left|v\right|}{m}-g*\alpha
\end{equation}

We can represent the second time derivative of velocity as jerk (rate of change of acceleration) as follows.

\begin{equation}\label{Equation 3.17}
\ddot{v}=j
\end{equation}

\subsubsection{Lateral Vehicle Dynamics}

\begin{figure}[htb]
	
	\centering
	\includegraphics[width=\textwidth]{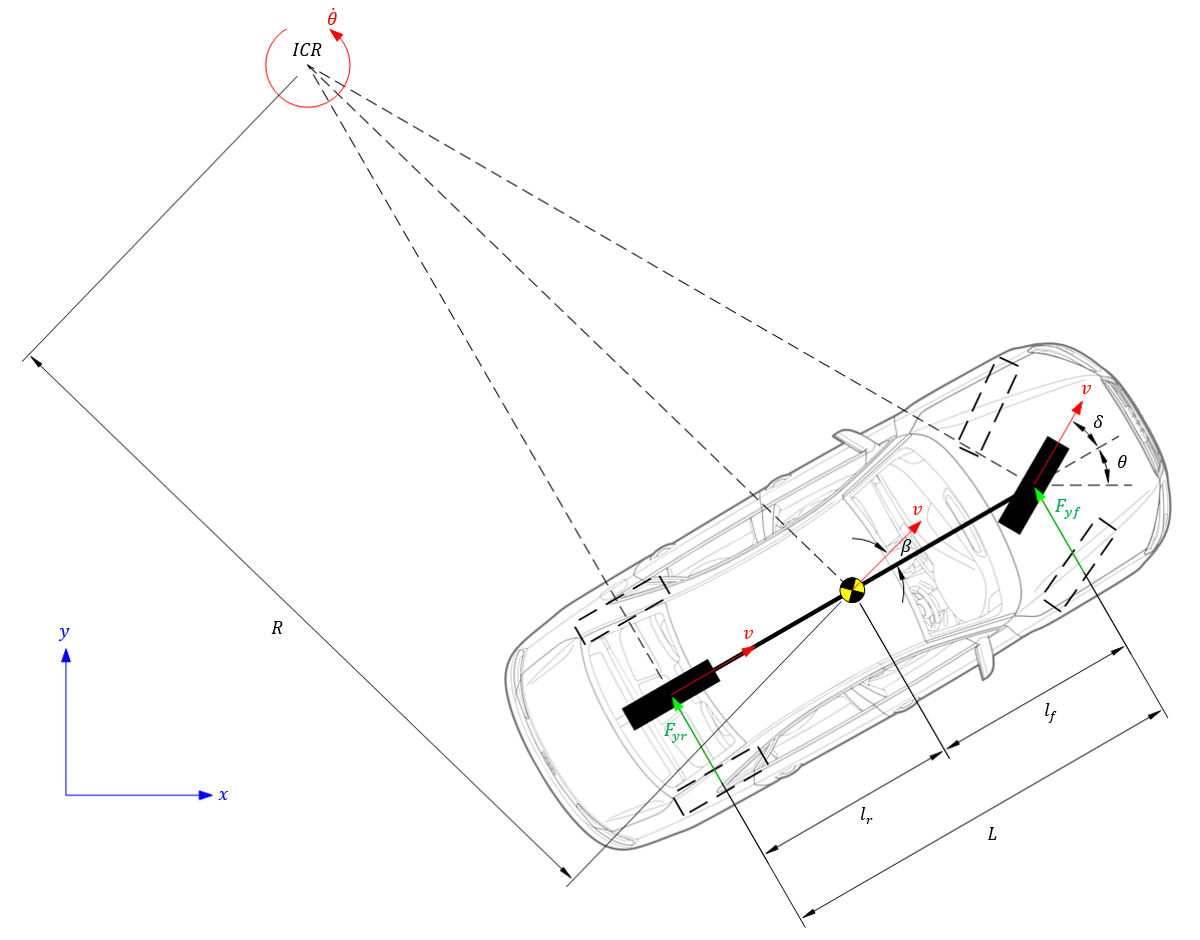}
	\caption{Lateral Vehicle Dynamics}
	\label{fig:Lateral Vehicle Dynamics}
	
\end{figure}

The free-body diagram of an autonomous vehicle along the lateral direction (denoted as $y$) is depicted in figure \ref{fig:Lateral Vehicle Dynamics}. The lateral forces considered include the inertial term $m*a_y$ and the front and rear tire forces $F_{yf}$ and $F_{yr}$, respectively. The torques considered include vehicle torque about instantaneous center of rotation $I_z*\ddot{\theta}$, and moments of front and rear tire forces $l_f*F_{yf}$ and $l_r*F_{yr}$, respectively (acting in opposite direction). Thus, applying Newton's second law of motion to the free-body diagram we get,

\begin{equation}\label{Equation 3.18}
m*a_y=F_{yf}+F_{yr}
\end{equation}

\begin{equation}\label{Equation 3.19}
I_z*\ddot{\theta}=l_f*F_{yf}-l_r*F_{yr}
\end{equation}

The linear lateral acceleration $\ddot{y}$ and centripetal acceleration $R*\dot{\theta}^2$ collectively contribute to the total lateral acceleration $a_y$ of the ego vehicle. Hence, we have,

\begin{equation}\label{Equation 3.20}
a_y=\ddot{y}+R*\dot{\theta}^2=\ddot{y}+v*\dot{\theta}
\end{equation}

\pagebreak

Substituting equation \ref{Equation 3.20} into equation \ref{Equation 3.18} we get,

\begin{equation}\label{Equation 3.21}
m*\left(\ddot{y}+v*\dot{\theta}\right)=F_{yf}+F_{yr}
\end{equation}

The linearized lateral tire forces $F_{yf}$ and $F_{yr}$, also known as cornering forces, can be modeled using front and rear tire slip angles $\alpha_f$ and $\alpha_r$, and cornering stiffness of front and rear tires $C_f$ and $C_r$ as follows.

\begin{equation}\label{Equation 3.22}
\left\{\begin{matrix}
F_{yf}=C_f*\alpha_f=C_f*\left(\delta-\beta-\frac{l_f*\dot{\theta}}{v}\right)\\ 
F_{yr}=C_r*\alpha_r=C_r*\left(-\beta+\frac{l_r*\dot{\theta}}{v}\right)
\end{matrix}\right.
\end{equation}

Substituting the values of linearized lateral tire forces $F_{yf}$ and $F_{yr}$ from equation \ref{Equation 3.22} into equations \ref{Equation 3.21} and \ref{Equation 3.19} respectively, and rearranging the terms we get,

\begin{equation}\label{Equation 3.23}
\ddot{y}=-\frac{\left(C_f+C_r\right)}{m}*\beta+\left(\frac{C_r*l_r-C_f*l_f}{m*v}-v\right)*\dot{\theta}+\frac{C_f}{m}*\delta
\end{equation}

\begin{equation}\label{Equation 3.24}
\ddot{\theta}=\frac{C_r*l_r-C_f*l_f}{I_z}*\beta-\frac{C_r*l_r^2+C_f*l_f^2}{I_z*v}*\dot{\theta}+\frac{C_f*l_f}{I_z}*\delta
\end{equation}

\subsubsection{Consolidated Dynamic Model}

Equations \ref{Equation 3.16}, \ref{Equation 3.17}, \ref{Equation 3.23} and \ref{Equation 3.24} represent the dynamic model of an autonomous vehicle. The consolidated continuous-time dynamic model of the autonomous vehicle can be therefore formulated as follows.

\begin{equation}\label{Equation 3.25}
\ddot{q}=\begin{bmatrix}\ddot{x}\\ \ddot{y}\\ \ddot{\theta}\\ \ddot{v}\end{bmatrix}=\begin{bmatrix}
r_{wheel}*\ddot{\theta}_{wheel}-\frac{C_\alpha*v^2}{m}-\frac{\hat{C}_{r,1}*\left|v\right|}{m}-g*\alpha\\ -\frac{\left(C_f+C_r\right)}{m}*\beta+\left(\frac{C_r*l_r-C_f*l_f}{m*v}-v\right)*\dot{\theta}+\frac{C_f}{m}*\delta\\ \frac{C_r*l_r-C_f*l_f}{I_z}*\beta-\frac{C_r*l_r^2+C_f*l_f^2}{I_z*v}*\dot{\theta}+\frac{C_f*l_f}{I_z}*\delta\\ j\end{bmatrix}
\end{equation}

Based on the formulation in equations \ref{Equation 3.25} and \ref{Equation 3.10}, we can formulate the discrete-time model of autonomous vehicle as follows.

\begin{equation}\label{Equation 3.26}
\left\{\begin{matrix}
x_{t+1}=x_t+\dot{x_t}*\Delta t+\ddot{x_t}*\frac{\Delta t^2}{2}\\
y_{t+1}=y_t+\dot{y_t}*\Delta t+\ddot{y_t}*\frac{\Delta t^2}{2}\\
\theta_{t+1}=\theta_t+\dot{\theta_t}*\Delta t+\ddot{\theta_t}*\frac{\Delta t^2}{2}\\
v_{t+1}=v_t+\dot{v_t}*\Delta t+\ddot{v_t}*\frac{\Delta t^2}{2}
\end{matrix}\right.
\end{equation}

Note that the equation \ref{Equation 3.26} is known as the \textit{state transition equation} (generally represented as $q_{t+1}=q_t+\dot{q_t}*\Delta t+\ddot{q_t}*\frac{\Delta t^2}{2}$) where $t$ in the subscript denotes current time instant and $t+1$ in the subscript denotes next time instant.

\section{Control Strategies}

Thus far, we have discussed in detail, all the background concepts related to control strategies for autonomous vehicles (or any other system for that matter).

In this section we present classical as well as some of the current state-of-the-art control strategies for autonomous vehicles. Some of these are pretty easy and intuitive, while others are not. We shall begin discussing simpler ones first and then introduce more complex strategies for vehicle control, but first, let us see the different control schemes.

\subsection{Control Schemes}

As stated in section \ref{Control System Architecture for Autonomous Vehicles}, the control system of an autonomous vehicle is split into longitudinal and lateral components. Depending on whether or not these components influence each other, the control schemes can be referred to as \textit{coupled} or \textit{de-coupled} control.

\subsubsection{Coupled Control}

Coupled control refers to a control scheme wherein the lateral and longitudinal controllers operate synergistically and have some form of influence over each other. Although this approach is very much realistic, the major difficulty in practical implementation lies in accurately determining the ``amount" of influence the controllers are supposed to have over one another considering the motion model and the operational design domain (ODD) of the ego vehicle.

In coupled-control scheme, the lateral controller may influence the longitudinal controller or vice-versa, or both may have an influence over each other. Let us consider each case independently with an example.

\paragraph{\textbf{Longitudinal Controller Influencing Lateral Controller}\\ \\}

In this case, the longitudinal controller is dominant or has a higher preference over the lateral controller. As a result, the longitudinal control action is computed independently, and this influences the lateral control action with an inverse relation (i.e. inverse proportionality). Since it is not a good idea to have a large steering angle at high speed, the controller regulates maximum steering limit inversely proportional to the vehicle speed.

\paragraph{\textbf{Lateral Controller Influencing Longitudinal Controller}\\ \\}

As opposed to the previous case, in this one, the lateral controller is dominant or has a higher preference over the longitudinal controller. As a result, the lateral control action is computed independently, and this influences the longitudinal control action with an inverse relation (i.e. inverse proportionality). Since it is a bad idea to have high speed with large steering angle, the controller regulates maximum speed limit inversely proportional to the steering angle.

\paragraph{\textbf{Longitudinal and Lateral Controllers Influencing Each Other}\\ \\}

In this case, both the controllers tend to influence each other depending upon the operating conditions. If it is more important to maintain high speeds, the longitudinal controller has a higher (more dominant) influence over the lateral controller and if the maneuver requires cornering at sharp turns, the lateral controller has a higher (more dominant) influence over the longitudinal controller. Generally, the operating conditions change over the coarse of duration and ``weights" need to be shifted in order to switch the dominance of lateral and longitudinal controllers.

\subsubsection{De-Coupled Control}

De-coupled control refers to a control scheme wherein the lateral and longitudinal controllers do not have any influence on each other, i.e. both the controllers operate independently disregarding the control actions generated by other.

Although this simplifies the control problem to a large extent, this type of control scheme does not seem much realistic considering the fact that lateral and longitudinal vehicle dynamics inherently affect each other. In other words, this control scheme is applicable to highly specific and uniform driving scenarios (i.e. limited ODD).

\subsection{Traditional Control} \label{Traditional Control}

Traditional control is one of the most widely used and reliable control strategy, especially when it comes to safety-critical systems such as autonomous vehicles.

There are primarily three objectives of a controller.

\begin{enumerate}
	
	\item \textbf{Stability:} Stability refers to the fact that a system should always produce a bounded output (response) when excited with a bounded input signal. This criterion for stability is popularly known as the BIBO criterion. This is especially significant for inherently unstable systems.
	
	\item \textbf{Tracking:} Tracking refers to the fact that a control system should be able to track the desired reference value (a.k.a. setpoint). In other words, the system response should be as close to the setpoint as possible, with minimal transient and/or steady state error (zero in ideal case).
	
	\item \textbf{Robustness:} Robustness refers to the fact that a control system should be able to exhibit stability and tracking despite external disturbances. In other words, the system must remain stable (i.e. meet the BIBO criterion) and its response should be as close to the setpoint as possible even when it is disturbed by external factors.
	
\end{enumerate}

Since traditional controllers can be tuned and analyzed to meet the objectives stated above, they become the primary choice of control strategies, especially for systems such as autonomous robots and vehicles.

Traditional control systems can either be open loop, which do not use any state feedback or closed loop, which do. However, when it comes to motion control of autonomous vehicles, a control problem where operating conditions are continuously changing, open loop control is not a possibility. Thus, for control of autonomous vehicles, closed loop control systems are implemented with an assumption that the state is directly measurable or can be estimated by implementing state observers. Such traditional controllers can be classified as \textit{model-free} and \textit{model-based} controllers.

\begin{enumerate}
	
	\item \textbf{Model-Free Controllers:} These type of controllers do not use any mathematical model of the system being controlled. They tend to take ``corrective" action based on the ``error" between setpoint and current state. These controllers are quite easy to implement since they do not require in-depth knowledge of the behavior of the system, however, are difficult to tune, do not guarantee optimal performance and perform satisfactorily under limited operating conditions. Common examples of these type of controllers include bang-bang controllers, PID controllers, intelligent PID controllers (iPIDs), etc.
	
	\item \textbf{Model-Based Controllers:} These type of controllers use some or the other type of mathematical model of the system being controlled. Depending upon the model complexity and the exact approach followed to generate the control commands, these can be further classified as follows.
	
	\begin{enumerate}
		
		\item \textbf{Kinematic Controllers:} These type of controllers use simplified motion models of the system and are based on the geometry and kinematics of the system (generally employing first order approximation). They assume no slip, no skip and often ignore internal or external forces acting on the system. Therefore, these type of controllers are often restricted to low-speed applications (where system dynamics can be approximated), but offer an advantage of low computational complexity (which is extremely significant in real-world implementation as opposed to theoretical formulation or simulation).
		
		\item \textbf{Dynamic Controllers:} These type of controllers use detailed motion models of the system and are based on system dynamics. They consider forces and torques acting on the system as well as any disturbances (incorporated in the dynamic model). Therefore, these type of controllers have an upper hand when compared to kinematic controllers due to the unrestricted ODD, but suffer higher computational complexity owing to the complex computations involving detailed models at each time step.
		
		\item \textbf{Model Predictive Controllers:} These type of controllers use linear or non-linear motion models of the system to predict its future states (up to a finite receding horizon) and determine the optimal control action by numerically solving a bounded optimization problem at each time step (essentially treating the control problem as an optimization problem, a strategy known as optimal control). The optimization problem is bounded since model predictive controllers explicitly handle motion constraints, which is another reason for their popularity. Although this sounds to be the perfect control strategy, it is to be noted that since model predictive controllers use complex motion models and additionally solve an online optimization problem at each time step, they are computationally expensive and may cause undesirable control latency (sometimes even of the order of a few seconds) if not implemented wisely.
		
	\end{enumerate}
	
\end{enumerate}

The upcoming sections discuss these traditional control strategies for autonomous vehicles in a greater detail.

\subsubsection{Bang-Bang Control}

The bang-bang controller is a simple binary controller. It is extremely easy to implement, which is the most significant (and perhaps the only) reason for adopting it. The control action $u$ is switched between two states $u_{max}$ and $u_{min}$ (analogous to on/off, high/low, true/false, set/reset, 1/0, etc.) based on whether the input signal $x$ is above or below the reference value $x_{ref}$.

\begin{equation*}
u=
\begin{cases}
u_{max} &; x < x_{ref}\\
u_{min} &; x > x_{ref}\\
\end{cases}
\end{equation*}

Note that a multi-step controller may be adopted to switch the control action based on more than just two cases. For example, a third control action of $0$ is also possible in case the error becomes exactly zero (i.e. $x = x_{ref}$); however, practically, this case will sustain only instantaneously, since even a small error value will overshoot the system response.

The bang-bang controller, in either of its two states, generates the same control action regardless of the error value. In other words, it does not account for the magnitude of error signal. It is, therefore, termed as ``unstable" due to its abrupt responses, and is recommended to be applied only to \textit{variable structure systems (VSS)} that allow \textit{sliding motion control (SMC)}.

Considering the final control elements (throttle, brake and steering) of an autonomous vehicle, it is safe to say that bang-bang controller may not be applied to control the lateral vehicle dynamics; the vehicle would constantly oscillate about the mean position, trying to minimize the cross-track error $e_{\Delta}$ and/or heading error $e_\psi$ by completely turning the steering wheel in one direction or the other, which would be not only uncomfortable but also dangerous and may even cause a rollover at higher velocities. Nonetheless, bang-bang controller may be applied to control the longitudinal vehicle dynamics by fully actuating the throttle and brakes based on the velocity error $e_v$ and while this may be safe, it may still result in exceeding the nominal jerk values, causing a rather uncomfortable ride. It is important to note at this point that abrupt control actions generated by a bang-bang controller can possibly cause lifetime reduction, if not immediate breakdown, of actuators, which is highly undesirable.

\paragraph{\textbf{Bang-Bang Control Implementation}\\ \\}

\begin{figure}[htb]
	
	\centering
	\includegraphics[width=\textwidth]{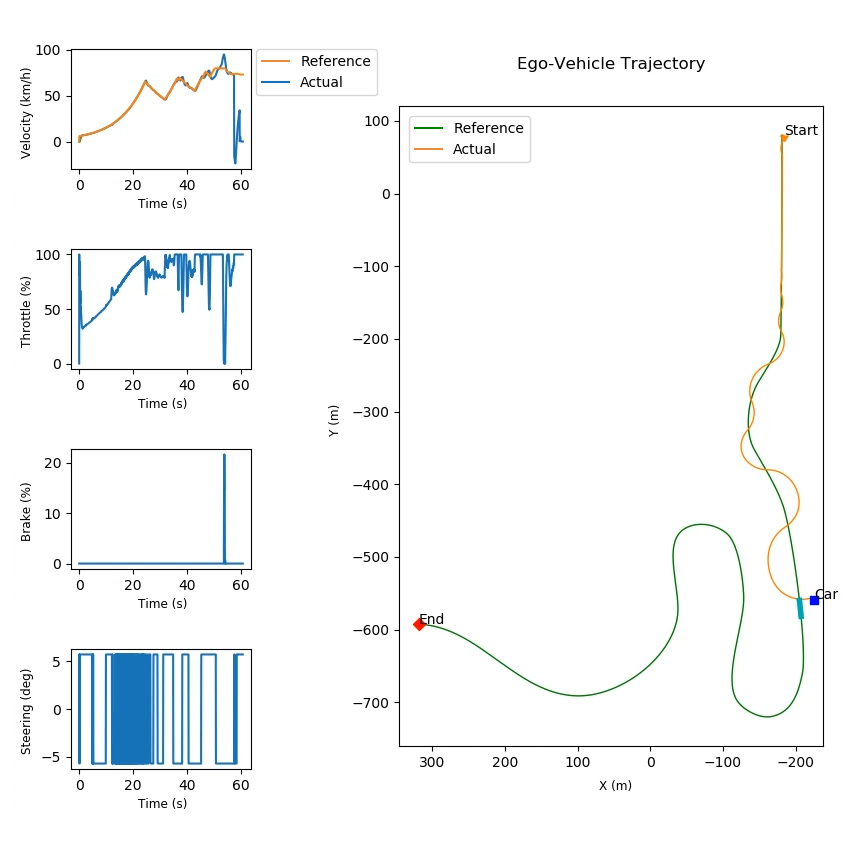}
	\caption{Bang-Bang Control Implementation}
	\label{fig:Bang-Bang Control Implementation}
	
\end{figure}

Implementation of bang-bang controller for lateral control of a simulated autonomous vehicle is illustrated in figure \ref{fig:Bang-Bang Control Implementation} (although, we talked about this not being a good idea). In order to have uniformity across all the implementations discussed in this chapter, a well-tuned PID controller was employed for longitudinal control of the ego vehicle.

For the purpose of this implementation, the reference trajectory to be followed by the ego vehicle (marked green in the trajectory plot) was discretized into waypoints to be tracked by the lateral controller, and each waypoint had an associated velocity setpoint to be tracked by the longitudinal controller.

The bang-bang controller response was alleviated by multiplying the control action with a limiting factor of $0.1$, thereby restricting the steering actuation limit $\delta \in [-7^{\circ},7^{\circ}]$. This allowed the bang-bang controller to track the waypoints at least for a while, after which, the ego vehicle went out of control and crashed!

The steering commands generated by the bang-bang controller are worth noting. It can be clearly seen how the controller switches back-and-forth between the two extremities, and although it was programmed to produce no control action (i.e. $0$) when the cross-track error was negligible, such an incident never occurred during the course of simulation. 

In reality, it is the abrupt control actions that quickly render the plant uncontrollable, which implies that the controller harms itself. Furthermore, it is to be noted that generating such abrupt control actions may not be possible practically, since the actuators have a finitely positive time-constant (especially steering actuation mechanism), and that doing so may cause serious damage to the actuation elements (especially at high frequencies, such as those observed between time-steps 15-25). This fact further adds to the controlability problem, rendering the bang-bang controller close to useless for controlling the lateral vehicle dynamics.

\subsubsection{PID Control}

From the previous discussion, it is evident that bang-bang controller is not a very promising option for vehicle control (especially lateral control, but also not so good for longitudinal control either) and better control strategies are required.

\paragraph{\textbf{P Controller}\\ \\}

The most intuitive upgrade from bang-bang controller is the proportional (P) controller. The P controller generates a control action $u$ proportional to the error signal $e$, and the proportionality constant that scales the control action is called gain of the P controller (generally denoted as $k_P$).

In continuous time, the P controller is represented as follows.

\begin{equation*}
u(t)=k_P*e(t)
\end{equation*}

In discrete time, the above equation takes the following form.

\begin{equation*}
u_{t+1}=k_P*e_t
\end{equation*}

It is extremely important to tune the gain value so as to obtain a desired system response. A low gain value would increase the settling time of the system drastically, whereas a high gain value would overshoot the system in the opposite direction. A moderate proportional gain would try to minimize the error, however, being an error-driven controller, it would still leave behind a significantly large steady-state error.

\paragraph{\textbf{PD Controller}\\ \\}

The PD controller can be thought of as a compound controller constituted of the proportional (P) and derivative (D) controllers.

The P component of the PD controller, as stated earlier, produces a ``surrogate" control action proportional to the error signal $e$. The proportional gain $k_P$ is deliberately set slightly higher so that the system would oscillate about the setpoint. The D component of the PD controller damps out the resulting oscillations by observing the temporal derivative (rate of change) of error and modifies the control action accordingly. The proportionality constant that scales the derivative control action is called gain of the D controller and is generally denoted as $k_D$. 

In continuous time, the PD controller is represented as follows.

\begin{equation*}
u(t)=k_P*e(t)+k_D*\frac{\mathrm{d}}{\mathrm{d}t}e(t)
\end{equation*}

In discrete time, the above equation takes the following form.

\begin{equation*}
u_{t+1}=k_P*e_t+k_D*\left[\frac{e_t-e_{t-1}}{\Delta t}\right]
\end{equation*}

The modified control action soothes the system response and reduces any overshoots. However, the system may still struggle in some cases, where its dynamics are disturbed due to physical interactions.

It is to be noted that the D controller is extremely susceptible to noise; even a small noise in the sensor readings can lead to miscalculation of error values resulting in larger or smaller derivatives, ultimately leading to instability. The gain $k_D$ must, therefore, be tuned wisely. Another remedy to this issue is to use a low-pass filter to reject the high-frequency components of the feedback signal, which are generally constituents of noise.

\paragraph{\textbf{PI Controller}\\ \\}

The PI controller can be thought of as a compound controller constituted of the proportional (P) and integral (I) controllers.

The P component of the PD controller, as stated earlier, produces a ``surrogate" control action proportional to the error signal $e$. The proportional gain $k_P$ is set to a moderate value so that the system tries really hard to converge to the setpoint. The I component of the PI controller modifies the control action based on the error accumulated over a certain time interval. The proportionality constant that scales the integral control action is called gain of the I controller and is generally denoted as $k_I$.

In continuous time, the PI controller is represented as follows.

\begin{equation*}
u(t)=k_P*e(t)+k_I*\int_{t_o}^{t} e(t)\:\mathrm{d}t
\end{equation*}

In discrete time, the above equation takes the following form.

\begin{equation*}
u_{t+1}=k_P*e_t+k_I*\sum_{i=t_o}^{t}e_i
\end{equation*}

Here, $t-t_o$ represents the temporal size of history buffer over which the error is integrated.

The modified control action forces the overall system response to get much closer to the setpoint value as the time proceeds. In other words, it helps the system better converge to the desired value. It is also worth mentioning that the I component is also effective in dealing with any systematic biases, such as inherent misalignments, disturbance forces, etc. However, with a simple PI controller implemented, the system may tend to overshoot every time a control action is executed.

It is to be noted that since the I controller acts on the accumulated error, which is a large value, its gain $k_I$ is usually set very low.

\paragraph{\textbf{PID Controller}\\ \\}

The proportional (P), integral (I) and derivative (D) controllers work in tandem to give rise to a much more efficient PID controller. The PID controller takes advantage of all the three primary controllers to generate a sophisticated control action that proportionally corrects the error, then dampens the resulting overshoots and reduces any steady-state error over the time.

In continuous time, the PID controller is represented as follows.

\begin{equation*}
u(t)=k_P*e(t)+k_I*\int_{t_o}^{t} e(t)\:\mathrm{d}t+k_D*\frac{\mathrm{d}}{\mathrm{d}t}e(t)
\end{equation*}

In discrete time, the above equation takes the following form.

\begin{equation*}
u_{t+1}=k_P*e_t+k_I*\sum_{i=t_o}^{t}e_i+k_D*\left[\frac{e_t-e_{t-1}}{\Delta t}\right]
\end{equation*}

As stated earlier, $t-t_o$ represents the temporal size of history buffer over which the error is integrated.

The gains of the designed PID controller need to be tuned manually at first. A general rule of thumb is to start by initializing the $k_I$ and $k_D$ values to zero and tune up the $k_P$ value until the system starts oscillating about the setpoint. The $k_D$ value is then tuned until the oscillations are damped out in most of the cases. Finally, the $k_I$ value is tuned to reduce any steady-state error. An optimizer algorithm (such as twiddle, gradient descent, etc.) can then be adopted to fine-tune the gains through recursive updates.

Characteristics of the controller gains (i.e. effect of gain amplification on closed loop system response) for the proportional, integral and derivative terms of the PID controller have been summarized in table \ref{Characteristics of PID Controller Gains}. This information is extremely useful for controller gain tuning in order to achieve the desired system response.

\begin{table}[h]
	\caption{Characteristics of PID Controller Gains}
	\label{Characteristics of PID Controller Gains}
	\centering
	\resizebox{\textwidth}{!}{%
		\begin{tabular}{|c|c|c|c|c|}
			\hline
			\textbf{\multirow{2}{*}{\begin{tabular}[c]{@{}c@{}}Controller\\ Gain\\ Amplification\end{tabular}}} & \multicolumn{4}{c|}{\textbf{Closed Loop System Response}}                                                                                                                                                                                                                      \\ \cline{2-5} 
			& \textbf{\begin{tabular}[c]{@{}c@{}}Rise Time\\ ($t_r$)\end{tabular}} & \textbf{\begin{tabular}[c]{@{}c@{}}Overshoot\\ ($M_p$)\end{tabular}} & \textbf{\begin{tabular}[c]{@{}c@{}}Settling Time\\ ($t_s$)\end{tabular}} & \textbf{\begin{tabular}[c]{@{}c@{}}Steady State Error\\ ($e_{ss}$)\end{tabular}} \\ \hline
			$k_P$                                                                                  & Decrease                                                    & Increase                                                    & Small Change                                                    & Decrease                                                                \\ \hline
			$k_I$                                                                                  & Decrease                                                    & Increase                                                    & Increase                                                        & Eliminate                                                               \\ \hline
			$k_D$                                                                                  & Small Change                                                & Decrease                                                    & Decrease                                                        & Small Change                                                            \\ \hline
		\end{tabular}
	}
\end{table}

In general, the P controller can be thought of as correcting the present error by generating a control action proportional to it. The I controller can be thought of as correcting any past error by generating a control action proportional to the error accumulated over time. Finally, the D controller can be thought of as correcting any future error by generating a control action proportional to the rate of change of error.

Note that the integral and derivative controllers cannot be employed alone, they can only assist the proportional controller. Same goes with the ID controller.

\paragraph{\textbf{PID Control Implementation}\\ \\}

\begin{figure}[htb]
	
	\centering
	\includegraphics[width=\textwidth]{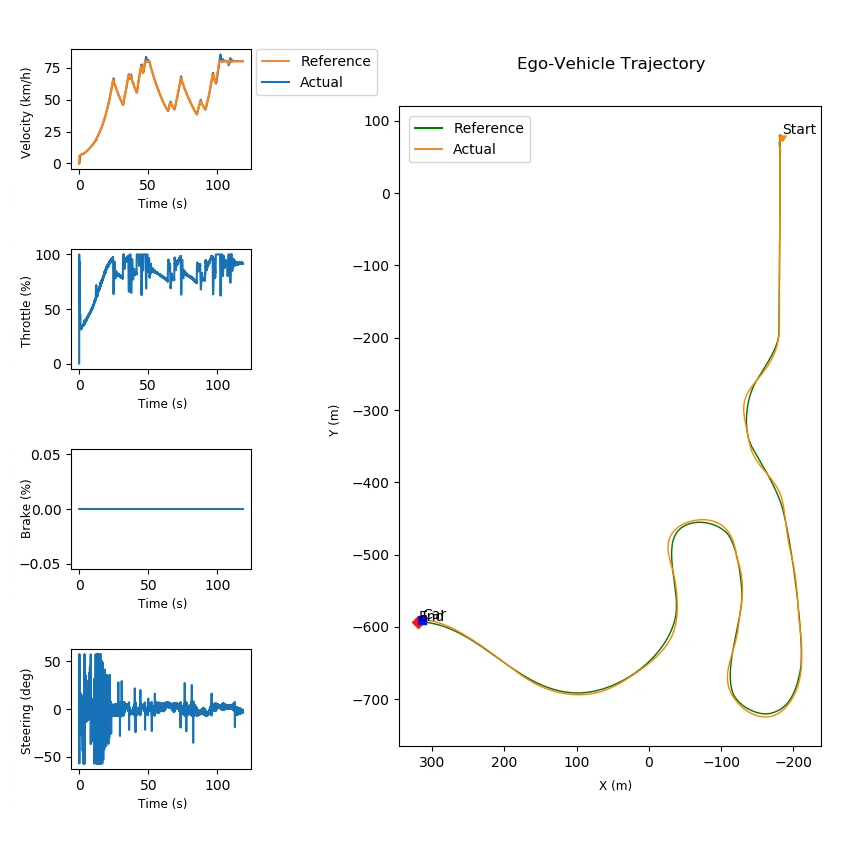}
	\caption{PID Control Implementation}
	\label{fig:PID Control Implementation}
	
\end{figure}

Implementation of PID controller for lateral control of a simulated autonomous vehicle is illustrated in figure \ref{fig:PID Control Implementation}.

The implementation as well employs a secondary PID controller for controlling the longitudinal vehicle dynamics. It is to be noted that the two controllers are decoupled and perform independently. In order to avoid confusion, the lateral controller shall be the point of interest for this topic, unless specified otherwise.

For the purpose of this implementation, the reference trajectory to be followed by the ego vehicle (marked green in the trajectory plot) was discretized into waypoints to be tracked by the lateral controller, and each waypoint had an associated velocity setpoint to be tracked by the longitudinal controller.

The PID controller was tuned in order to obtain best possible results for tracking the entire reference trajectory (which included both straight as well as curved segments) at varying longitudinal velocities. It is to be noted that although the controller does not compromise with comfort or safety, the tracking accuracy is quite limited even after recursive gain tuning, especially at higher road curvatures.

It can be therefore concluded that a standalone PID controller is more of a ``satisfactory" solution to the problem of controlling lateral vehicle dynamics, and that advanced strategies (such as gain scheduling) and/or better controllers are a more attractive choice for this task.

\subsubsection{Geometric Control}

Geometric control refers to the notion of computing the control commands for the ego vehicle purely using the ``geometry" of the vehicle kinematics and reference trajectory.

Although this type of control strategy is applicable to both lateral as well as longitudinal control, PID controller is a better choice when it comes to longitudinal control as compared to geometric controllers since it operates directly on the velocity error. For lateral control though, geometric controllers are a great choice since PID controller for lateral vehicle control needs to be tuned for specific velocity ranges, beyond which it is either too sluggish or too reactive.

There are two highly popular geometric path tracking controllers for lateral motion control of autonomous vehicles viz. Pure Pursuit and Stanley controllers.

\paragraph{\textbf{Pure Pursuit Controller}\\ \\}

Pure Pursuit controller is a geometric trajectory tracking controller that uses a \textit{look-ahead point} on the reference trajectory at a fixed distance ahead of the ego vehicle in order to determine the cross-track error $e_\Delta$. The controller then decides the steering angle command in order to minimize this cross-track error and ultimately try to reach the look-ahead point. However, since the look-ahead point is at a fixed distance ahead of the ego vehicle, the vehicle is constantly in pursuit of getting to that point and hence this controller is called Pure Pursuit controller.

\begin{figure}[htb]
	
	\centering
	\includegraphics[width=\textwidth]{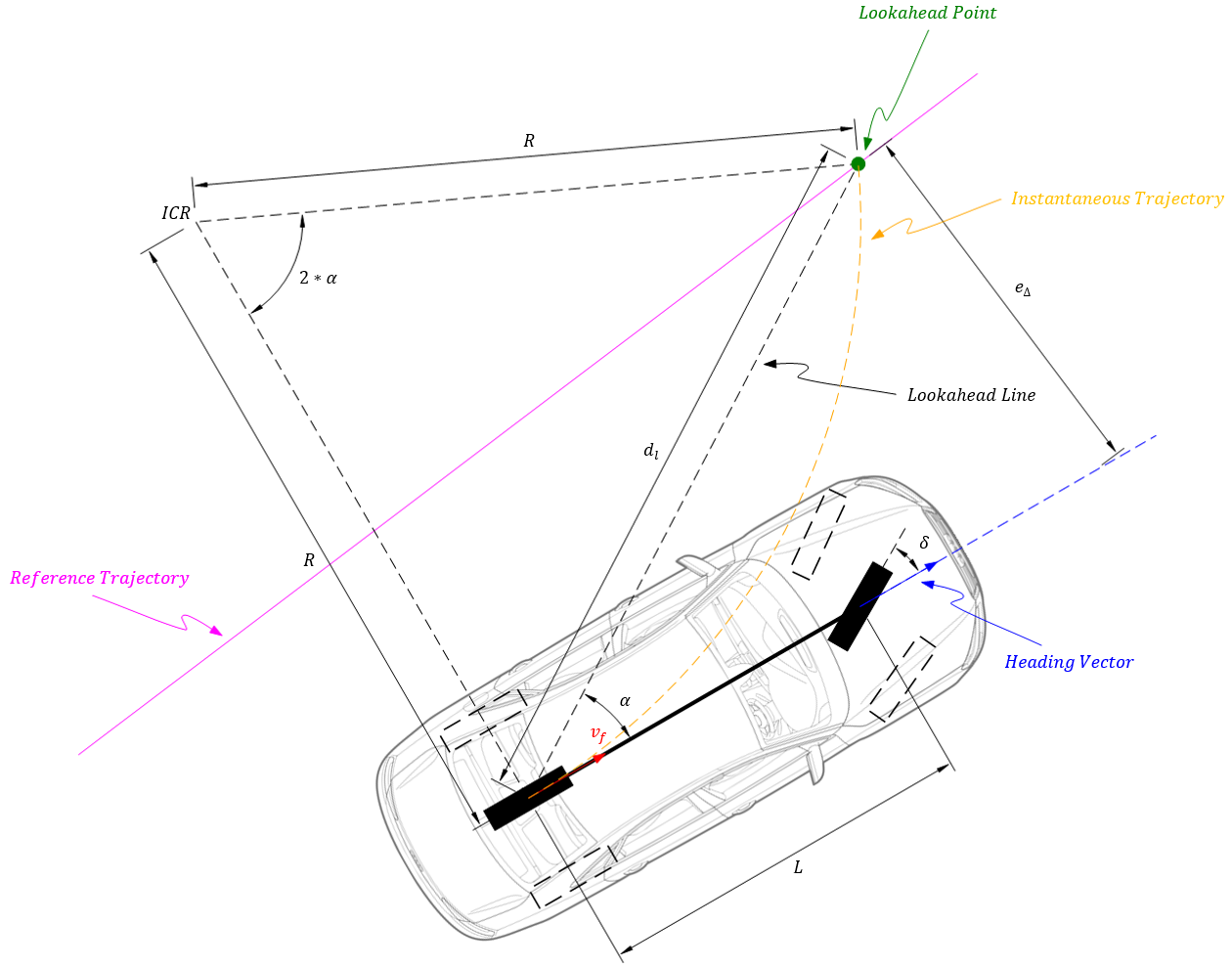}
	\caption{Pure Pursuit Control}
	\label{fig:Pure Pursuit Control}
	
\end{figure}

If we consider the center of the rear axle of the ego vehicle as the frame of reference, we have the geometric relations as shown in figure \ref{fig:Pure Pursuit Control}. Here, the ego vehicle wheelbase is $L$, forward velocity of the ego vehicle is $v_f$, steering angle of the ego vehicle is $\delta$, distance between lookahead point on reference trajectory and the center of the rear axle of the ego vehicle is the lookahead distance $d_l$, the angle between vehicle heading and the lookahead line is $\alpha$. Since the vehicle is a rigid body with forward steerable wheels, for non-zero $v_f$ and $\delta$, the vehicle follows a circular path of radius $R$ (corresponding to curvature $\kappa$) about the instantaneous center of rotation (ICR).

With reference to figure \ref{fig:Pure Pursuit Control}, from the law of sines, we have,

\begin{equation*}
\frac{d_l}{sin\left(2*\alpha\right)}=\frac{R}{sin\left(\frac{\pi}{2}-\alpha\right)}
\end{equation*}

\begin{equation*}
\therefore \frac{d_l}{2*sin(\alpha)*cos(\alpha)}=\frac{R}{cos\left(\alpha\right)}
\end{equation*}

\begin{equation*}
\therefore \frac{d_l}{sin(\alpha)}=2*R
\end{equation*}

\begin{equation} \label{Equation 3.27}
\therefore \kappa=\frac{1}{R}=\frac{2*sin(\alpha)}{d_l}
\end{equation}

\begin{equation} \label{Equation 3.28}
\because sin(\alpha)=\frac{e_\Delta}{d_l} \Rightarrow \kappa=\frac{2}{d_l^2}*e_\Delta \Rightarrow \kappa\propto e_\Delta
\end{equation}

It can be seen from equation \ref{Equation 3.28} that curvature of the instantaneous trajectory $\kappa$ is directly proportional to the cross-track error $e_\Delta$. Thus, as the cross-track error increases, so does the trajectory curvature, thereby bringing the vehicle back towards the reference trajectory aggressively. Note that the term $\frac{2}{d_l^2}$ can be thought of as a proportionality gain, which can be tuned based on the lookahead distance parameter.

With reference to figure \ref{fig:Pure Pursuit Control}, the steering angle command $\delta$ can be computed using the following relation.

\begin{equation} \label{Equation 3.29}
tan(\delta)=\frac{L}{R}=L*\kappa
\end{equation}

Substituting the value of $\kappa$ from equation \ref{Equation 3.27} in equation \ref{Equation 3.29} and solving for $\delta$ we get,

\begin{equation} \label{Equation 3.30}
\delta=tan^{-1}\left(\frac{2*L*sin(\alpha)}{d_l}\right)
\end{equation}

Equation \ref{Equation 3.30} presents the de-coupled scheme of Pure Pursuit controller (i.e. the steering law is independent of vehicle velocity). As a result, if the controller is tuned for low speed, it will be dangerously aggressive at higher speeds while if tuned for high speed, the controller will be too sluggish at lower speeds. One potentially simple improvement would be to vary the lookahead distance $d_l$ proportional to the vehicle velocity $v_f$ using $k_v$ as the proportionality constant (this constant/gain will act as the tuning parameter for Pure Pursuit controller).

\begin{equation} \label{Equation 3.31}
d_l=k_v*v_f
\end{equation}

Substituting the value of $d_l$ from equation \ref{Equation 3.31} in equation \ref{Equation 3.30} we get the complete coupled Pure Pursuit control law formulation as follows.

\begin{equation} \label{Equation 3.32}
\delta=tan^{-1}\left(\frac{2*L*sin(\alpha)}{k_v*v_f}\right)\:;\delta \in \left[-\delta_{max},\delta_{max}\right]
\end{equation}

In summary, it can be seen that Pure Pursuit controller acts as a geometric proportional controller of steering angle $\delta$ operating on the cross-track error $e_\Delta$ while observing the steering actuation limits.

\paragraph{\textbf{Pure Pursuit Control Implementation}\\ \\}

\begin{figure}[htb]
	
	\centering
	\includegraphics[width=\textwidth]{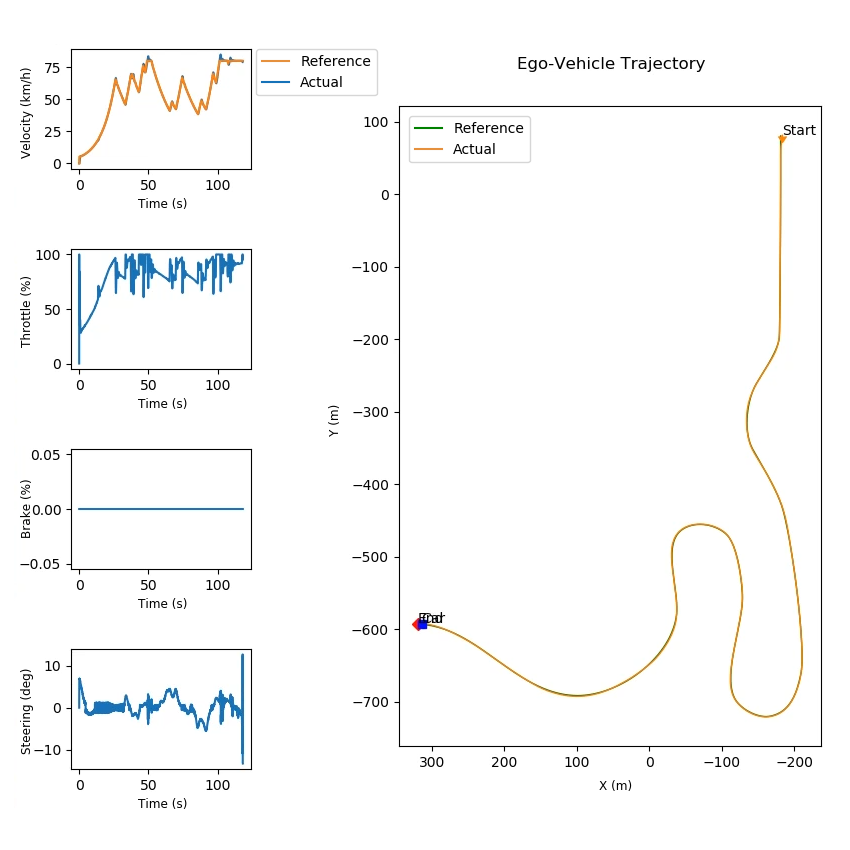}
	\caption{Pure Pursuit Control Implementation}
	\label{fig:Pure Pursuit Control Implementation}
	
\end{figure}

Implementation of Pure Pursuit controller for lateral control of a simulated autonomous vehicle is illustrated in figure \ref{fig:Pure Pursuit Control Implementation}. The implementation as well employs a PID controller for controlling the longitudinal vehicle dynamics. It is to be noted that this implementation uses coupled control scheme wherein the steering action is influenced by the vehicle velocity following an inverse relation.

For the purpose of this implementation, the reference trajectory to be followed by the ego vehicle (marked green in the trajectory plot) was discretized into waypoints to be tracked by the lateral controller, and each waypoint had an associated velocity setpoint to be tracked by the longitudinal controller.

The lateral controller parameter $k_v$ was tuned in order to obtain best possible results for tracking the entire reference trajectory (which included both straight as well as curved segments) at varying forward velocities. The controller was very well able to track the prescribed trajectory with a promising level of precision. Additionally, the control actions generated during the entire course of simulation lied within a short span $\delta \in [-15^{\circ}, 15^{\circ}]$, which is an indication of the controller having a good command over the system.

Nonetheless, a green highlight (representing the reference trajectory) can be seen in the neighborhood of actual trajectory followed by the ego vehicle, which indicates that the tracking was not ``perfect". This may be partially due to the fact that the Pure Pursuit control law has a single tunable parameter $d_l$ having an approximated first order relation with $v_f$. However, a more convincing reason for potential errors in trajectory tracking, when it comes to Pure Pursuit controller, is that it generates steering action regardless of heading error of the vehicle, making it difficult to actually align the vehicle precisely along the reference trajectory.

Furthermore, being a kinematic controller, Pure Pursuit controller disregards actual system dynamics, and as a result compromises with trajectory tracking accuracy. Now, although this may not affect a vehicle driving under nominal conditions, a racing vehicle, for example, would experience serious trajectory deviations under rigorous driving conditions.

Pure Pursuit controller also has a pretty non-intuitive issue associated with it, which is generally undiscovered during infinitely continuous or looped trajectory tracking. Particularly, towards the end of a finite reference trajectory, the Pure Pursuit controller generates erratic steering actions owing to the fact that the \textit{look-ahead point} it uses in order to deduce the steering control action is no longer available, since the further waypoints are not really defined. This may lead to undesired stoppage of the ego vehicle or even cause it to wander off of the trajectory in the final few seconds of the mission.

In conclusion, Pure Pursuit controller is really good for regulating the lateral dynamics of a vehicle operating under nominal driving conditions, and that there is room for further improvement.

\paragraph{\textbf{Stanley Controller}\\ \\}

Stanley controller is a geometric trajectory tracking controller developed by Stanford University's ``Stanford Racing Team" for their autonomous vehicle ``Stanley" at the DARPA Grand Challenge (2005). As opposed to Pure Pursuit controller (discussed earlier), which uses only the cross-track error to determine the steering action, Stanley controller uses both heading as well as cross-track errors to determine the same. The cross-track error $e_\Delta$ is defined with respect to a closest point on the reference trajectory (as opposed to a lookahead point in case of Pure Pursuit controller) whereas the heading error $e_\psi$ is defined using the vehicle heading relative to the reference trajectory.

\begin{figure}[htb]
	
	\centering
	\includegraphics[width=\textwidth]{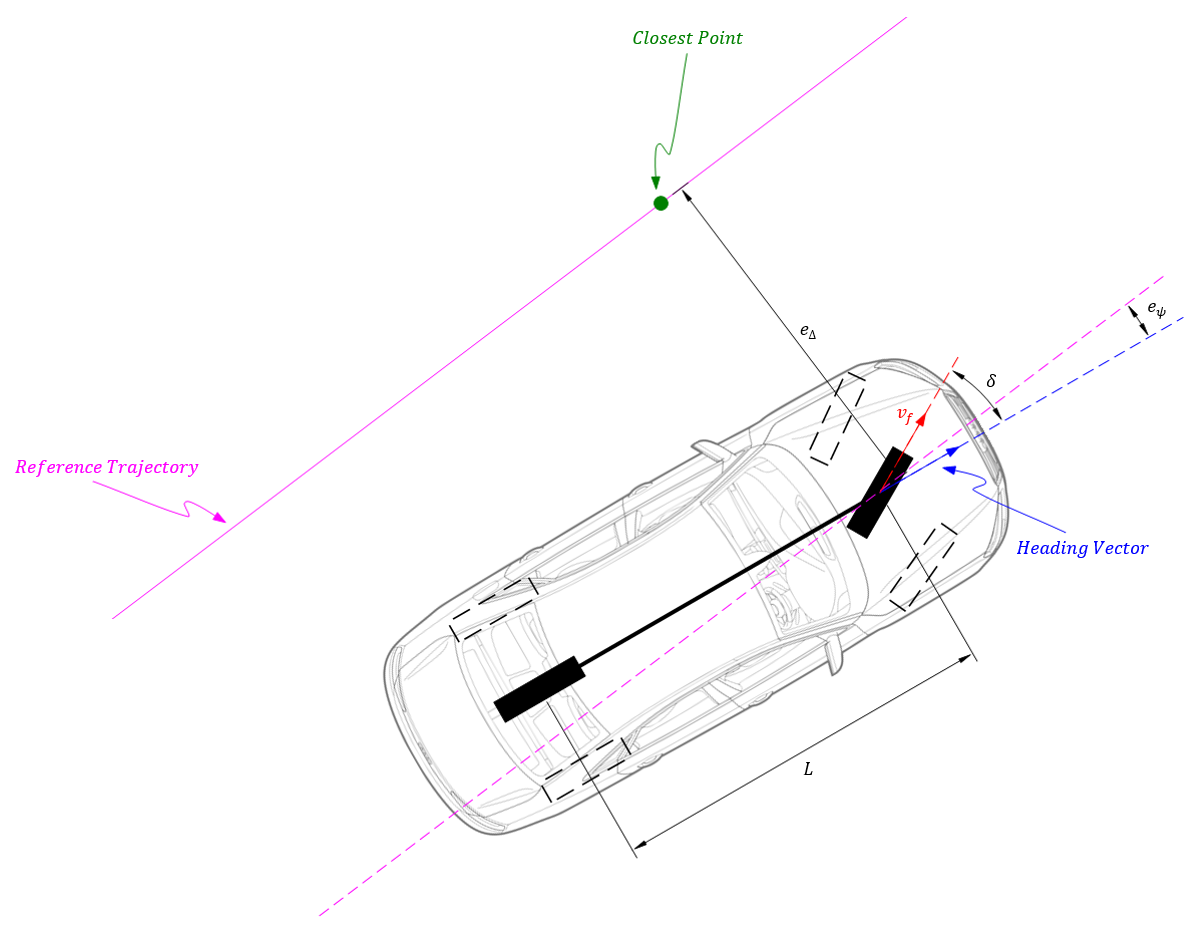}
	\caption{Stanley Control}
	\label{fig:Stanley Control}
	
\end{figure}

If we consider the center of the front axle of the ego vehicle as the frame of reference, we have the geometric relations as shown in figure \ref{fig:Stanley Control}. Here, the ego vehicle wheelbase is $L$, forward velocity of the ego vehicle is $v_f$, and steering angle of the ego vehicle is $\delta$.

Stanley controller uses both heading as well as cross-track errors to determine the steering command. Furthermore, the steering angle generated by this (or any other) method must observe the steering actuation limits. Stanley control law, therefore, is essentially defined to meet the following three requirements.

\begin{enumerate}
	
	\item \textbf{Heading Error Correction:} To correct the heading error $e_\psi$ by producing a steering control action $\delta$ proportional (or equal) to it, such that vehicle heading aligns with the desired heading.
	
	\begin{equation} \label{Equation 3.33}
	\delta=e_\psi
	\end{equation}
	
	\item \textbf{Cross-Track Error Correction:} To correct the cross-track error $e_\Delta$ by producing a steering control action $\delta$ directly proportional to it and inversely proportional to the vehicle velocity $v_f$ in order to achieve coupled-control. Moreover, the effect for large cross-track errors can be limited by using an inverse tangent function.
	
	\begin{equation} \label{Equation 3.34}
	\delta=tan^{-1}\left(\frac{k_\Delta*e_\Delta}{v_f}\right)
	\end{equation}
	
	It is to be noted that at this stage of the formulation, the inverse relation between steering angle and vehicle speed can cause numerical instability in control actions.
	
	At lower speeds, the denominator becomes small, thus causing the steering command to shoot to higher values, which is undesirable considering human comfort. Hence, an extra softening coefficient $k_s$ may be used in the denominator as an additive term in order to keep the steering commands smaller for smoother steering actions.
	
	On the contrary, at higher velocities, the denominator becomes large making the steering commands small in order to avoid large lateral accelerations. However, even these small steering actions might be high in some cases, causing high lateral accelerations. Hence, an extra damping coefficient $k_d$ may be used in order to dampen the steering action proportional to vehicle velocity.
	
	\begin{equation} \label{Equation 3.35}
	\delta=tan^{-1}\left(\frac{k_\Delta*e_\Delta}{k_s+k_d*v_f}\right)
	\end{equation}
	
	\item \textbf{Clipping Control Action:} To continuously observe the steering actuation limits $[-\delta_{max}, \delta_{max}]$ and clip the steering command within these bounds.
	
	\begin{equation} \label{Equation 3.36}
	\delta \in \left[-\delta_{max},\delta_{max}\right]
	\end{equation}
	
\end{enumerate}

Using equations \ref{Equation 3.33}, \ref{Equation 3.35} and \ref{Equation 3.36} we can formulate the complete Stanley control law as follows.

\begin{equation} \label{Equation 3.37}
\delta=e_\psi+tan^{-1}\left(\frac{k_\Delta*e_\Delta}{k_s+k_d*v_f}\right)\:;\delta \in \left[-\delta_{max},\delta_{max}\right]
\end{equation}

In summary, it can be seen that Stanley controller acts as a geometric proportional controller of steering angle $\delta$ operating on the heading error $e_\psi$ as well as the cross-track error $e_\Delta$, while observing the steering actuation limits.

\paragraph{\textbf{Stanley Control Implementation}\\ \\}

\begin{figure}[htb]
	
	\centering
	\includegraphics[width=\textwidth]{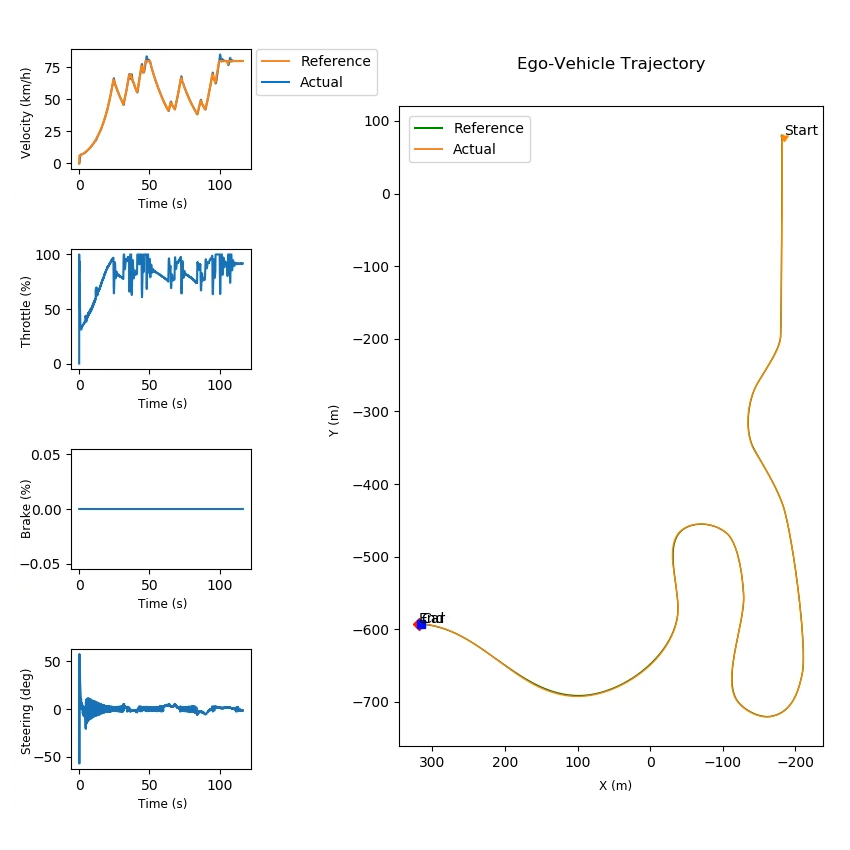}
	\caption{Stanley Control Implementation}
	\label{fig:Stanley Control Implementation}
	
\end{figure}

Implementation of Stanley controller for lateral control of a simulated autonomous vehicle is illustrated in figure \ref{fig:Stanley Control Implementation}. The implementation as well employs a PID controller for controlling the longitudinal vehicle dynamics. It is to be noted that this implementation uses coupled control scheme wherein the steering action is influenced by the vehicle velocity following an inverse relation.

For the purpose of this implementation, the reference trajectory to be followed by the ego vehicle (marked green in the trajectory plot) was discretized into waypoints to be tracked by the lateral controller, and each waypoint had an associated velocity setpoint to be tracked by the longitudinal controller.

The lateral controller parameters $k_\Delta$, $k_s$ and $k_d$ were tuned in order to obtain best possible results for tracking the entire reference trajectory (which included both straight as well as curved segments) at varying forward velocities. The controller was able to track the prescribed trajectory with a promising level of precision.

Nonetheless, a slight green highlight (representing the reference trajectory) can be occasionally spotted in the neighborhood of actual trajectory followed by the ego vehicle, which indicates that the tracking was sub-optimal or near-perfect. This may be due to the first order approximations involved in the formulation. Furthermore, being a kinematic geometric controller, Stanley controller disregards actual system dynamics, and as a result compromises with trajectory tracking accuracy, especially under rigorous driving conditions.

Stanley controller uses the closest waypoint to compute the cross-track error, owing to which, it may become highly reactive at times. Since this implementation assumed static waypoints defined along a racetrack (i.e. in the global frame), and the ego vehicle was spawned towards one side of the track, the initial cross-track error was quite high. This made the controller react erratically, thereby generating steering commands of over $50^{\circ}$ in either direction. However, this is not a significant problem practically, since the local trajectory is recursively planned as a finite set of waypoints originating from the vehicle coordinate system, thereby ensuring that the immediately next waypoint is not too far from the vehicle.

In conclusion, Stanley controller is one of the best for regulating lateral dynamics of a vehicle operating under nominal driving conditions, especially considering the computational complexity at which it offers such accuracy and robustness.

\subsubsection{Model Predictive Control}

Model predictive control (MPC) is a type of optimal control strategy that basically treats the control task as a constrained or bounded optimization problem. Model predictive controller predicts the future states of the system (ego vehicle) up to a certain prediction horizon using the motion model and then solves an online optimization problem considering the constraints (or control bounds) in order to select the optimal set of control inputs by minimizing a cost function such that the future state(s) of the ego vehicle closely align with the goal state (as required for trajectory tracking).

\begin{figure}[htb]
	
	\centering
	\includegraphics[width=\textwidth]{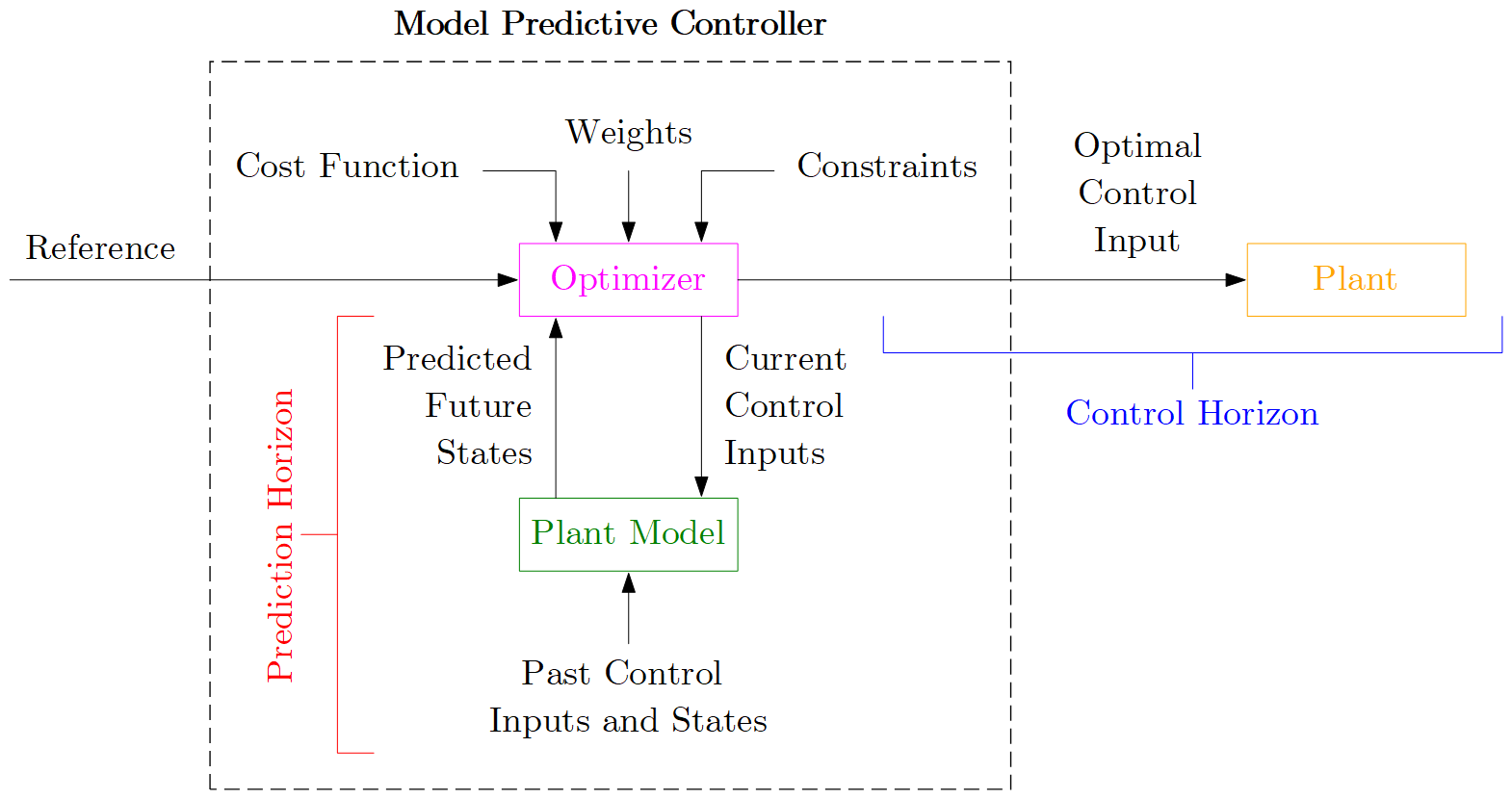}
	\caption{Model Predictive Control Architecture}
	\label{fig:Model Predictive Control Architecture}
	
\end{figure}

In other words, given the current state and the reference trajectory to follow, MPC involves simulating different control inputs (without actually applying them to the system), predicting the resulting future states (in form of a predicted trajectory) using motion model up to a certain prediction horizon, selecting the optimal set of control inputs corresponding to minimal cost trajectory (considering constraints) at each step in time, and applying the very first set of optimal control inputs (up to a certain control horizon) to the ego vehicle, discarding the rest. With the updated state, we again repeat the same algorithm to compute a new optimal predicted trajectory up to the prediction horizon. In that sense, we are computing optimal control inputs over a constantly moving prediction horizon. Thus, this approach is also known as \textit{receding horizon control}.

There are two main reasons for following the receding horizon approach in model predictive control.

\begin{enumerate}
	
	\item The motion model is only an approximate representation of the actual vehicle dynamics and despite our best efforts, it won't match the real world exactly. Hence, once we apply the optimal control input to the ego vehicle, our actual trajectory may not be exactly same as the trajectory we predicted. It is therefore extremely crucial that we constantly re-evaluate our optimal control actions in a receding horizon manner so that we do not build up a large error between the predicted and actual vehicle state.
	
	\item Beyond a specific prediction horizon, the environment will change enough that it won't make sense to predict any further into the future. It is therefore a better idea to restrict the prediction horizon to a finite value and follow the receding horizon approach to predict using the updated vehicle state at each time interval.
	
\end{enumerate}

Model predictive control is extremely popular in autonomous vehicles for the following reasons.

\begin{itemize}
	
	\item It can handle multi-input multi-output (MIMO) systems that have cross-interactions between the inputs and outputs, which very well suits for the vehicle control problem.
	
	\item It can consider constraints or bounds to compute optimal control actions. The constraints are often imposed due to actuation limits, comfort bounds and safety considerations, violating which may potentially lead to uncontrollable, uncomfortable or unsafe scenarios, respectively.
	
	\item It has a future preview capability similar to feedforward control (i.e. it can incorporate future reference information into the control problem to improve controller performance for smoother state transitioning).
	
	\item It can handle control latency. Since MPC uses system model for making an informed prediction, we can incorporate any control latency (time difference between application of control input and actual actuation) into the system model thereby enabling the controller to adopt to the latency.
	
\end{itemize}

Following are some of the practical considerations for implementing MPC for motion control of autonomous vehicles.

\begin{itemize}
	
	\item \textbf{Motion Model:} For predicting the future states of the ego vehicle, we can use kinematic or dynamic motion models as described in section \ref{Mathetical Modeling}. Note that depending upon the problem statement, one may choose simpler models, the exact same models, or even complex models. However, often, in most implementations (for nominal driving conditions) it is suggested to use kinematic models since they offer a good balance between simplicity and accuracy.
	
	\item \textbf{MPC Design Parameters:}
	
	\begin{itemize}
		
		\item \textbf{Sample Time ($T_s$):} Sample time determines the rate at which the control loop is executed. If it is too large, the controller response will be sluggish and may not be able to correct the system fast enough leading to accidents. Also, high sample time makes it difficult to approximate a continuous reference trajectory by discrete paths. This is known as \textit{discretization error}. On the other hand, if the sample time is too small, the controller might become highly reactive (sometimes over-reactive) and may lead to a uncomfortable and/or unsafe ride. Also, smaller the sample time, more is the computational complexity since an entire control loop is supposed to be executed within the time interval (including online optimization). Thus, a proper sample time must be chosen such that the controller is neither sluggish, nor over-reactive but can quickly respond to disturbances or setpoint changes. It is recommended to have sample time $T_s$ of 5 to 10 percent of the rise time $t_r$ of open loop step response of the system.
		
		\begin{equation*}
		0.05*t_r\leqslant T_s \leqslant 0.1*t_r
		\end{equation*}
		
		\item \textbf{Prediction Horizon ($p$):} Prediction horizon is the number of time steps in the future over which state predictions are made by the model predictive controller. If it is too small, the controller may not be able to take the necessary control actions sufficiently in advance, making it ``too late" in some situations. On the other hand, a large prediction horizon will make the controller predict too long into the future making it a wasteful effort, since a major part of (almost entire) predicted trajectory will be discarded in each control loop. Thus, a proper prediction horizon must be chosen such that it covers significant dynamics of the system and at the same time, is not excessively high. It is recommended to determine the prediction horizon $p$ depending upon the sample time $T_s$ and the settling time $t_s$ of open loop step response of the system (at 2\% steady state error criterion).
		
		\begin{equation*}
		\frac{t_s}{T_s}\leqslant p \leqslant 1.5*\frac{t_s}{T_s}
		\end{equation*}
		
		\item \textbf{Control Horizon ($m$):} Control horizon is the number of time steps in the future for which the optimal control actions are computed by the optimizer. If it is too short, the optimizer may not return the best possible control action(s). On the other hand, if the control horizon is longer, model predictive controller can make better predictions of future states and thus the optimizer can find best possible solutions for control actions. One can also make the control horizon equal to the prediction horizon, however, note that usually only a first couple of control actions have a significant effect on the predicted states. Thus, excessively larger control horizon only increases the computational complexity without being much of a help. Therefore, a general rule of thumb is to have a control horizon $m$ of 10 to 20 percent of the prediction horizon $p$.
		
		\begin{equation*}
		0.1*p\leqslant m \leqslant 0.2*p
		\end{equation*}
		
		\item \textbf{Constraints:} Model predictive controller can incorporate constraints on control inputs (and their derivatives) and the vehicle state (or predicted output). These can be either hard constraints (which cannot be violated under any circumstances) or soft constraints (which can be violated with minimum necessary amount). It is recommended to have hard constraints for control inputs thereby accounting for actuation limits. However, for outputs and time derivatives (time rate of change) of control inputs, it is not a good idea to have hard constraints as these constraints may conflict with each other (since none can be violated) and lead to an infeasible solution for the optimization problem. It is therefore recommended to have soft constraints for outputs and time derivatives of control inputs, which may be occasionally violated (if required). Note that in order to keep the violation of soft constraints small, it is minimized by the optimizer.
		
		\item \textbf{Weights:} Model predictive controller has to achieve multiple goals simultaneously (which may compete/conflict with each other) such as minimizing the error between current state and reference while limiting the rate of chance of control inputs (and obeying other constraints). In order to ensure a balanced performance between these competing goals, it is a good idea to weigh the goals in order of importance or criticality. For example, since it is more important to track the ego vehicle pose than it's velocity (minor variations in velocity do not affect much) one may assign higher weight to pose tracking as compared to velocity tracking. This will cause the optimizer to give more weightage to the vehicle pose as compared to it's velocity (similar to how hard constraints are more weighted as opposed to soft constraints).
		
	\end{itemize}
	
	\item \textbf{Cost Function ($J$):} The exact choice of cost function depends very much upon the specific problem statement requirements, and is up to the control engineer. To state an example, for longitudinal control, one may use velocity error $e_v$ or distance to goal $e_\chi$ as cost, while for lateral control, one may use the cross-track error $e_{\Delta}$ and/or heading error $e_\psi$ as the cost. Practically, it is advised to use quadratic (or higher degree) cost functions as opposed to linear ones since they penalize more for deviations from reference, and tend to converge the optimization faster. Furthermore, it is a good idea to associate cost not only for the error from desired reference $e$, but also for the amount of change in control inputs $\Delta u$ between each time step so that comfort and safety are maintained. For that, one may choose a cost function that considers the weighted squared sums of predicted errors and control input increments as follows.
	
	\begin{equation*}
	J=\sum_{i=1}^{p}w_e*e_{t+i}^2+\sum_{i=0}^{p-1}w_{\Delta u}*\Delta u_{t+i}^2;\:\text{where}\begin{cases}t\rightarrow\text{present}\\t+i\rightarrow\text{future}\end{cases}
	\end{equation*}
	
	\item \textbf{Optimization:} The main purpose of optimization is to choose the optimal set of control inputs $u_{optimal}$ (from a set of all plausible controls $u$) corresponding to the predicted trajectory with lowest cost (whilst considering current state $x$, reference state $x_{ref}$ and constraints). As a result, there is no restriction on the optimizer to be used for this purpose and the choice is left to the control engineer who should consider the cost function $J$ as well as the time complexity required for solving the optimization problem. Practically, it is a good idea to set a tolerance till which minimization optimization is to be carried out, beyond which, optimization should be terminated and optimal set of control inputs corresponding to the predicted trajectory with lowest cost should be returned. A general optimization function might look something like the following.
	\begin{equation*}
	u_{optimal}=argmin\left(J\:|\:u,x,x_{ref},\left\langle\text{optimizer}\right\rangle,\left\langle\text{constraints}\right\rangle,\left\langle\text{tolerance}\right\rangle\right)
	\end{equation*}
	
\end{itemize}

Following are some of the variants of MPC depending upon the nature of system (plant model), constraints and cost function.

\begin{enumerate}
	
	\item \textbf{Linear Time-Invariant MPC:} If the plant model and constraints are linear, and the cost function is quadratic, it gives rise to \textit{convex optimization problem} where there is a single global minima and a variety of numerical methods exist to solve such optimization problems. Thus, we can use linear time invariant model predictive controller to control such systems.
	
	\item \textbf{Linear Time-Variant MPC:} If the plant model is non-linear, we may need to linearize it at different operation points (varying in time) thereby calling for linear time-variant model predictive controller.
	
	\begin{itemize}
		
		\item \textbf{Adaptive MPC:} If the structure of the optimization problem remains the same across all operating conditions (i.e. states and constraints do not change with operating conditions), we can simply approximate the non-linear plant model to a linear model (linearization) and use an adaptive model predictive controller which updates the plant model recursively as the operating conditions change.
		
		\item \textbf{Gain-Scheduled MPC:} If the plant model is non-linear and the states and/or constraints change with the operating conditions, we need to use gain-scheduled model predictive controller. In this approach, we perform offline linearization at operating points of interest and for each operating point, we design an independent linear model predictive controller, considering the states and constraints for that particular operating point. We then select the suitable linear MPC for a specific range of operating conditions and switch between these linear model predictive controllers as operating conditions change.
		
	\end{itemize}
	
	\item \textbf{Non-Linear MPC:} If it is not possible/recommended to linearize the plant model, we need to use non-linear model predictive controller. Although this method is the most powerful one since it uses most accurate representation of the plant model, it is by far the most challenging one to solve in real-time since non-linear constraints and cost function give rise to \textit{non-convex optimization problem} with multiple local optima. It is quite difficult to find the global optimum and getting stuck in the local optima is possible. This approach is therefore highly complex in terms of computation and its efficiency depends upon the non-linear solver used for optimization.
	
\end{enumerate}

A general rule of thumb for selecting from the variants of MPC is to start simple and go complex if and only if it is necessary. Thus, linear time-invariant or traditional MPC and adaptive MPC are the two most commonly used approaches for autonomous vehicle control under nominal driving conditions.

\paragraph{\textbf{Model Predictive Control Implementation}\\ \\}

\begin{figure}[htb]
	
	\centering
	\includegraphics[width=\textwidth]{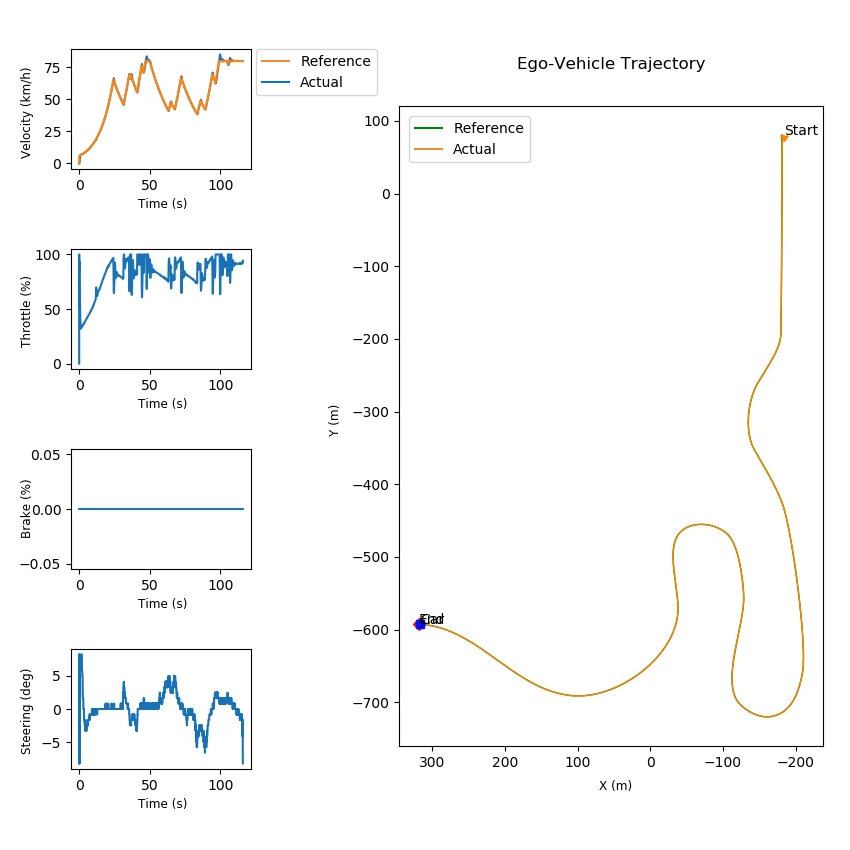}
	\caption{Model Predictive Control Implementation}
	\label{fig:Model Predictive Control Implementation}
	
\end{figure}

Implementation of model predictive controller for lateral control of a simulated autonomous vehicle is illustrated in figure \ref{fig:Model Predictive Control Implementation}. The implementation as well employs a PID controller for controlling the longitudinal vehicle dynamics.

For the purpose of this implementation, the reference trajectory to be followed by the ego vehicle (marked green in the trajectory plot) was discretized into waypoints to be tracked by the lateral controller, and each waypoint had an associated velocity setpoint to be tracked by the longitudinal controller.

A simple kinematic bicycle model also worked really well for tracking the entire reference trajectory (which included both straight as well as curved segments) at varying longitudinal velocities. The model predictive controller was able to track the prescribed trajectory with a promising level of precision and accuracy. Additionally, the steering control commands generated during the entire course of simulation lied well within the span $\delta \in [-10^{\circ}, 10^{\circ}]$, which is an indication of the controller being able to generate optimal control actions at right time instants.

One of the most significant drawback of MPC is its heavy requirement of computational resources. However, this can be easily counteracted by limiting the prediction and control horizons to moderately-small and small magnitudes, formulating simpler cost functions, using less detailed motion models or setting a higher tolerance for optimization convergence. Although this may lead to sub-optimal solutions, it reduces computational overhead to a great extent, thereby ensuring real-time execution of the control algorithm.

In conclusion, model predictive control can be regarded as the best control strategy for both simplistic as well as rigorous driving behaviors, provided a sufficiently powerful computational resource is available online.

\subsection{Learning Based Control}

As stated in section \ref{Traditional Control}, traditional controllers require in-depth knowledge of the process flow involved in designing and tuning them. Additionally, in case of model based controllers, the physical parameters of the system, along with its kinematic and/or dynamic models need to be known before designing a suitable controller. To add to the problem, most traditional controllers are scenario-specific and do not adapt to varying operating conditions; they need to be re-tuned in case of a scenario-change, which makes them quite inconvenient to work with. Lastly, most of the advanced traditional controllers are computationally expensive, which induces a time lag derived error in the processing pipeline.

Recently, learning based control strategies have started blooming, especially end-to-end learning. Such controllers hardly require any system-level knowledge and are much easier to implement. The system engineer may not need to implement the perception, planning and control sub-systems, since the AI agent learns to perform these operations implicitly; it is rather difficult to exactly tell which part of the neural network acts as the perception sub-system, which acts as planning sub-system and which acts as the control sub-system. Although the ``learning" process is computationally quite expensive, this step may be performed offline for a single time, after which, the trained model may be incorporated into the system architecture of the autonomous vehicle for a real-time implementation. Another advantage of such controllers is their ability to generalize across a range of similar scenarios, which makes them fit for minor deviations in the driving conditions. All in all, learning based control schemes seem to be a tempting alternative for the traditional ones. However, there is still a long way in achieving this goal as learning based controllers are somewhat unreliable; if presented with an unseen scenario, they can generate erratic control actions based on the learned features. Constant research is being carried out in this field to understand the way these controllers learn, and guiding them to learn the most appropriate features.

The learning based control strategies can be broadly classified into two categories, viz. \textit{imitation learning} and \textit{reinforcement learning}. While the former is a supervised learning strategy, the later is more of a self-learning technique. The upcoming sections discuss these strategies in a greater detail.

It is to be noted that, apart from end-to-end control, hybrid control strategies are also possible, meaning some aspects of traditional controllers (such as gains of PID controller or system model for MPC) can be ``learned" through recursive training (using imitation/reinforcement learning), thereby improving the performance of the controller.

\subsubsection{Imitation Learning Based Control}

Imitation learning is adopted to train an AI agent using a set of labeled data. The agent learns to map the features to the labels, while minimizing the error in its predictions. This technique produces fairly good results even at an early stage of the training process. However, achieving perfection using this technique requires a larger dataset, a sufficiently prolonged training duration and lot of hyperparameter tuning.

\begin{figure}[htb]
	
	\centering
	\includegraphics[width=\textwidth]{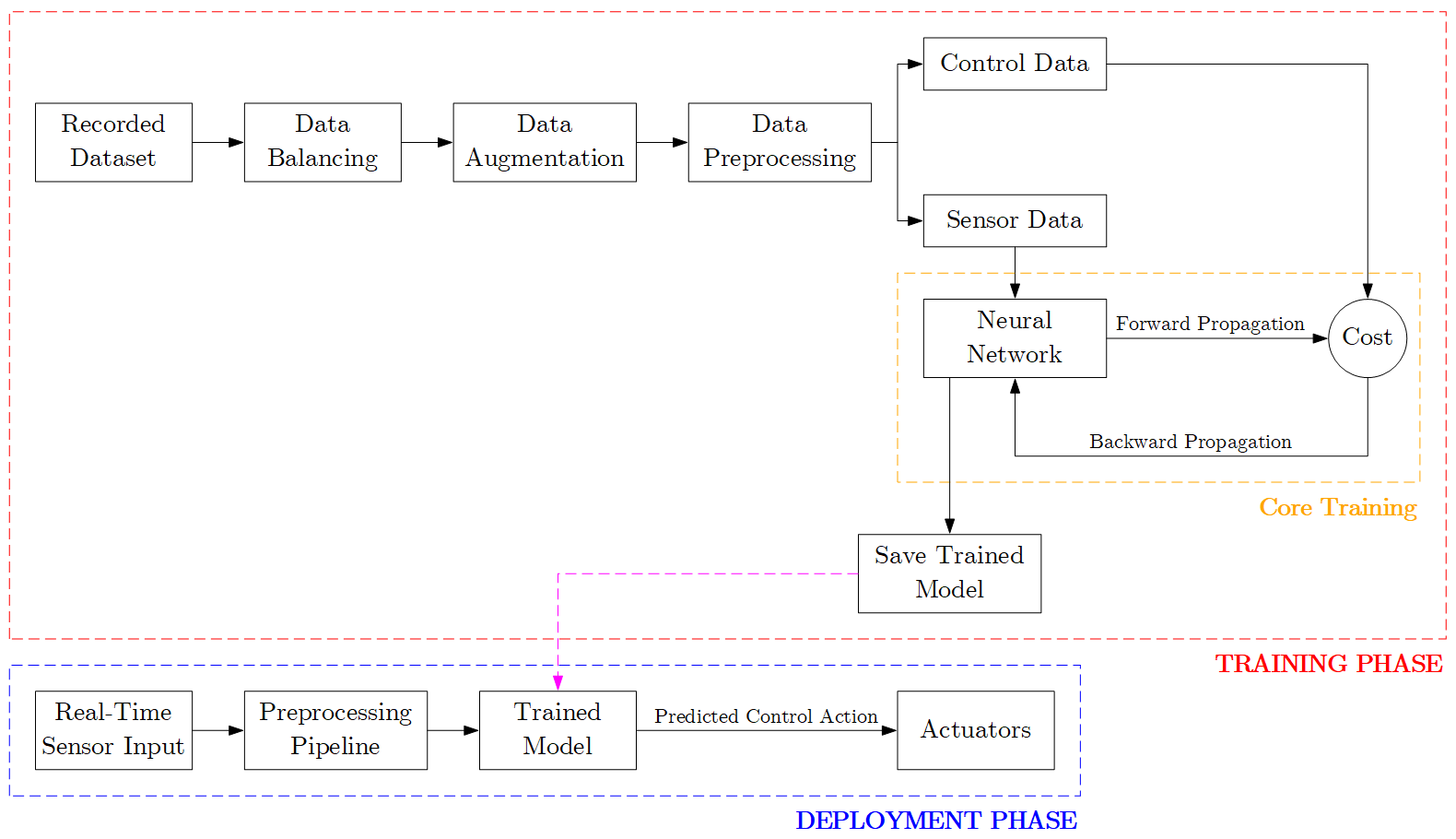}
	\caption{Imitation Learning Architecture}
	\label{fig:Imitation Learning Architecture}
	
\end{figure}

We shall consider \textit{behavioral cloning}, an exemplar implementation of this technique, as a case study. Behavioral cloning is concerned with cloning the driving behavior of an agent (a human driver or a traditional controller) by learning specific features from the training dataset provided. Figure \ref{fig:Imitation Learning Architecture} illustrates the training and deployment phases of imitation learning, which are discussed in the following sections.

\paragraph{\textbf{Training Phase\\ \\}}

During the training phase, the ego vehicle is driven (either manually or using a traditional controller) in a scenario resembling one in the deployment phase, and timestamped sensor readings (images, point cloud, etc.) along with the control input measurements (throttle, brake and steering) are recorded in a driving log. The collected dataset is then balanced to remove any inherent prejudice after which, it is augmented and preprocessed, thereby imparting robustness to the model and helping it generalize to a range of similar scenarios. Practically, the augmentation and preprocessing steps are performed on-the-go during the training process using a \textit{batch generator}. Finally, a deep neural network is ``trained" to directly predict the control inputs based on sensor readings.

It is to be noted that, practically, the collected dataset influences the training process more than anything else. Other important factors affecting the training, in order of importance, include neural network architecture (coarse-tuning is sufficient in most cases), augmentation and preprocessing techniques used and hyperparameter values chosen (fine-tuning may be required based on the application). It is also completely possible that one training pipeline works well on one dataset, while the other works well on a different one.

Training a neural network end-to-end in order to predict the control actions required to drive an autonomous vehicle based on the sensory inputs is essentially modeling the entire system as a linear combination of input \textit{features} to predict the output \textit{labels}. The training process can be elucidated as follows.

First a suitable neural network architecture is defined; this architecture may be iteratively modified based on experimental results. The trainable parameters of the model include \textit{weights} and \textit{biases}. Generally, the weights are initialized to a small non-zero random value, whereas the biases are initialized to zero.

The network then performs, what is called, a \textit{forward propagation} step. Here, each neuron of every layer of the neural network calculates a weighted average of the activated input vector $a$ it receives from the previous layer, based on its current weight vector $w_i$ and adds a bias term $b_i$ to it. For the first layer, however, the input vector $x$ is considered in place of $a^{\left[l-1\right]}$. The result of this operation $z_i$ is then passed through a non-linear activation function $g(z)$, to produce an activated output $a_i$.

\begin{equation*}
z_i^{\left[l\right]}=w_i^T\cdot a^{\left[l-1\right]}+b_i
\end{equation*}

\begin{equation*}
a_i^{\left[l\right]}=g^{\left[l\right]}(z_i^{\left[l\right]})
\end{equation*}

Note that $l$ here, denotes the $l^{th}$ layer of the neural network and $i$ denotes the $i^{th}$ neuron of that layer.

Practically, however, these operations are not performed sequentially for each and every neuron. Instead, \textit{vectorization} is adopted, which groups the similar variables and vectors into distinct matrices and performs all the matrix manipulations at once, thereby reducing the computational complexity of the operation. The activated output vector of last layer is termed as predicted output $\hat{y}$ of the neural network.

The predicted output $\hat{y}$ is then used to compute the loss $L$ of the forward propagation, based on the loss function defined. Intuitively, loss is basically a measure of incorrectness (i.e. error) of the predicted output w.r.t. the labeled output.

Generally, for a regression problem like the one we are talking about, the most commonly adopted loss function is the \textit{mean squared error (MSE)}. It is defined as follows.

\begin{equation*}
L(\hat{y},y)=\frac{1}{n}\sum_{i=1}^{n}\left\|y_i-\hat{y_i}\right \|^2
\end{equation*}

For classification problems, though, the \textit{cross-entropy} loss function is a favorable choice. For a dataset with $M$ different classes, it is defined as follows.

\begin{equation*}
L(\hat{y},y)=\begin{cases}
-(y_i*log(\hat{y_i})+(1-y_i)*log(1-\hat{y_i})) &; M=2\\ 
-\sum_{i=1}^{M}y_i*log(\hat{y_i}) &; M>2
\end{cases}
\end{equation*}

The equations defined above are used to compute loss for each training example. These individual losses $L$ over $m$ training examples can be used to evaluate an overall cost $J$ using the following relation.

\begin{equation*}
J(w,b) = \frac{1}{m}\sum_{i=1}^{m}L(\hat{y^{(i)}},y^{(i)})
\end{equation*}

Note that $w$ and $b$ in the above equation denote vectorized form of weights and biases respectively, across the entire training set (or a mini-batch).

The cost is then minimized by recursively updating the weights and biases using an optimizer. This step is known as \textit{back propagation}.

The most traditional optimizer is the \textit{gradient decent} algorithm. In its very rudimentary form, the algorithm updates trainable parameters (i.e. weights and biases) by subtracting from them their respective gradients $dw$ and $db$ w.r.t. the cost function $J$, scaled by the \textit{learning rate} hyperparameter $\alpha$.

\begin{equation*}
\begin{cases}
w^{\left[l\right]}=w^{\left[l\right]}-\alpha*dw^{\left[l\right]}\\
b^{\left[l\right]}=b^{\left[l\right]}-\alpha*db^{\left[l\right]}
\end{cases}
\end{equation*}

Where, $dw$ and $db$ are computed as follows.

\begin{equation*}
\begin{cases}
dw^{\left[l\right]}=\frac{\partial J}{\partial w^{\left[l\right]}}=dz^{\left[l\right]}\cdot a^{\left[l-1\right]}\\ 
db^{\left[l\right]}=\frac{\partial J}{\partial b^{\left[l\right]}}=dz^{\left[l\right]}
\end{cases}
\end{equation*}

Here, $dz$ and $da$, interdependent on each other, can be computed using the following relations.

\begin{equation*}
\begin{cases}
dz^{\left[l\right]}=da^{\left[l\right]}*g^{\left[l\right]'}(z^{\left[l\right]})\\ 
da^{\left[l-1\right]}=w^{\left[l\right]^T}\cdot dz^{\left[l\right]}
\end{cases}
\end{equation*}

However, the current state-of-the-art optimizer adopted by most domain experts is the \textit{Adam} optimizer. The algorithm commences by computing the gradients $dw$ and $db$ using the gradient descent method and then computes the exponential moving averages of the gradients $V$ and their squares $S$, similar to \textit{Momentum} and \textit{RMSprop} algorithms respectively.

\begin{equation*}
\begin{cases}
V_{dw}=\beta_1*V_{dw}+(1-\beta_1)*dw\\
V_{db}=\beta_1*V_{db}+(1-\beta_1)*db
\end{cases}
\end{equation*}

\begin{equation*}
\begin{cases}
S_{dw}=\beta_2*S_{dw}+(1-\beta_2)*dw^2\\
S_{db}=\beta_2*S_{db}+(1-\beta_2)*db^2
\end{cases}
\end{equation*}

The bias correction equations for the exponential moving averages after $t$ iterations are given as follows.

\begin{equation*}
\begin{cases}
\widehat V_{dw}=\frac{V_{dw}}{(1-\beta_1^t)}\\
\widehat V_{db}=\frac{V_{db}}{(1-\beta_1^t)}
\end{cases}
\end{equation*}

\begin{equation*}
\begin{cases}
\widehat S_{dw}=\frac{S_{dw}}{(1-\beta_2^t)}\\
\widehat S_{db}=\frac{S_{db}}{(1-\beta_2^t)}
\end{cases}
\end{equation*}

The final update equations for the training parameters are therefore as given below.

\begin{equation*}
\begin{cases}
w=w-\alpha*\frac{\widehat V_{dw}}{\sqrt{\widehat S_{dw}}+\epsilon}\\
b=b-\alpha*\frac{\widehat V_{db}}{\sqrt{\widehat S_{db}}+\epsilon}
\end{cases}
\end{equation*}

Note that $\epsilon$ in the above equations is a small factor that prevents division by zero. The authors of Adam optimizer propose the default values of $0.9$ for $\beta_1$, $0.999$ for $\beta_2$ and $10^{-8}$ for $\epsilon$.They show empirically that Adam works well in practice and compares favorably to other adaptive learning based algorithms.

\paragraph{\textbf{Deployment Phase}\\ \\}

During the deployment phase, the trained model is incorporated into the control architecture of the ego vehicle to predict (forward propagation step alone) the same set of control inputs (throttle, brake and steering) based on measurements (images, point cloud, etc.) from the same set of sensors as during the training phase, in real time.

It is worth mentioning that the pipeline adopted for the deployment phase should have minimum latency in order to ensure real-time performance.

\paragraph{\textbf{Imitation Learning Based Control Implementation}\\ \\}

\begin{figure}[htb]
	
	\centering
	\includegraphics[width=\textwidth]{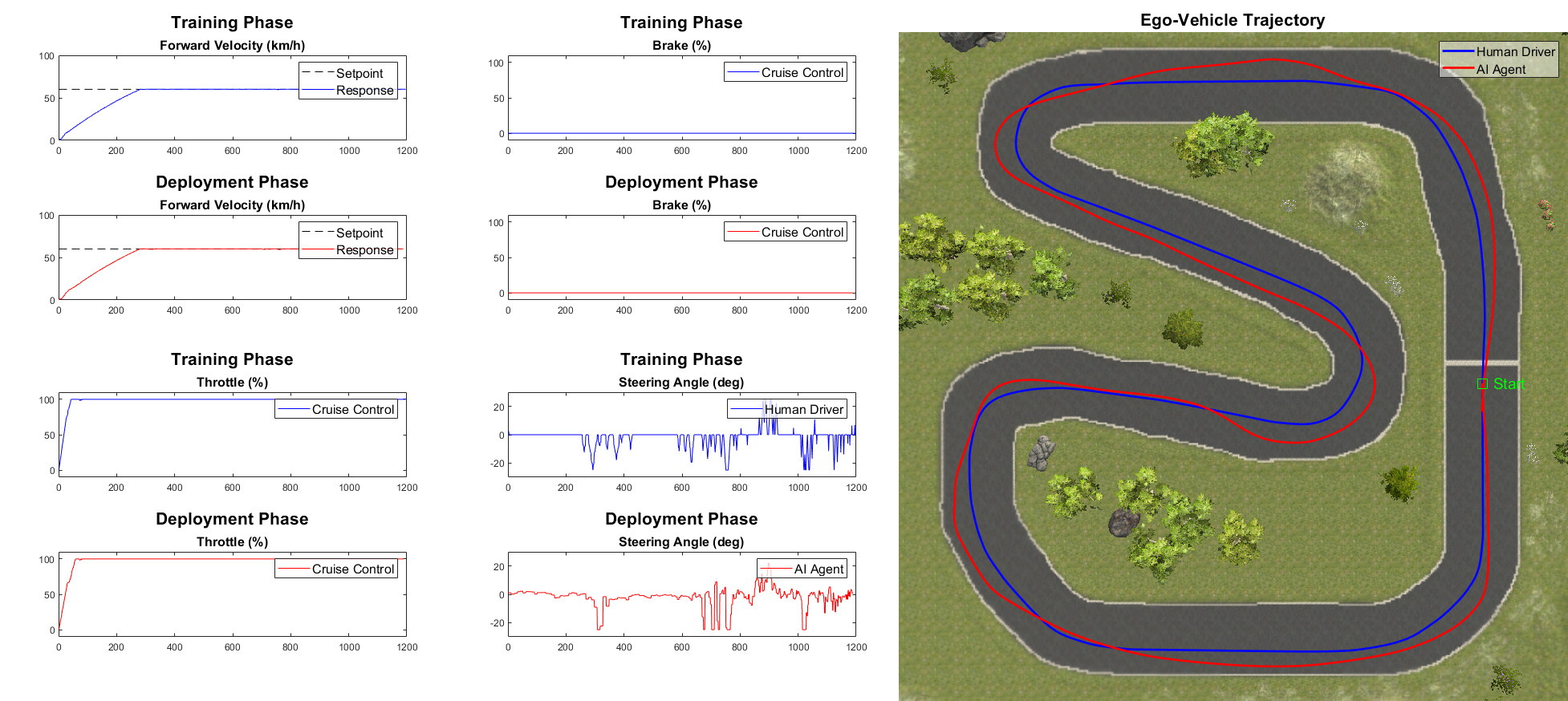}
	\caption{Imitation Learning Based Control Implementation}
	\label{fig:Imitation Learning Based Control Implementation}
	
\end{figure}

Implementation of imitation learning strategy for lateral control of a simulated autonomous vehicle is illustrated in figure \ref{fig:Imitation Learning Based Control Implementation}. Specifically, the longitudinal vehicle dynamics are controlled using a pre-tuned PID controller, while the lateral vehicle dynamics are controlled using a neural-network trained end-to-end.

The ego vehicle was manually driven for 20 laps, wherein a front-facing virtual camera mounted on the vehicle hood captured still frames of the road ahead. The steering angle corresponding to each frame was also recorded. This comprised the labeled dataset for training a deep neural network to mimic the driving skills of the human driver in an end-to-end manner (a.k.a. behavioral cloning). The trained neural network model was saved at the end of training phase.

The trained model was then deployed onto the ego vehicle, which was let to drive autonomously on the same track so as to validate the autonomy. The AI agent was successful in cloning most of the human driving behavior and exhibited $100\%$ autonomy, meaning it could complete several laps of the track autonomously. The lap time achieved by AI agent trained using imitation learning was 52.9 seconds, as compared to $\sim48.3$ seconds (statistical average) achieved by human driver.

The greatest advantage of adopting imitation learning for control of autonomous vehicles is its simplistic implementation and its ability to train fairly quickly (as compared to reinforcement learning). The neural network simply ``learns" to drive using the dataset provided, without the need of any complex formulations governing the input-output relations of the system. However, for the very same reason, this approach is highly susceptible to mislabeled or biased data, in which case, the neural network may learn incorrectly (this can be thought of as a bad teacher imparting erroneous knowledge to the students).

In conclusion, imitation learning is a convincing control strategy; however, advancements are required before achieving sufficient accuracy and reliability.

\subsubsection{Reinforcement Learning Based Control}

Imitation learning simply teaches the learner to mimic the trainer's behavior and the learner may never surpass performance of its trainer. This is where reinforcement learning comes into picture.

\begin{figure}[htb]
	
	\centering
	\includegraphics[width=\textwidth]{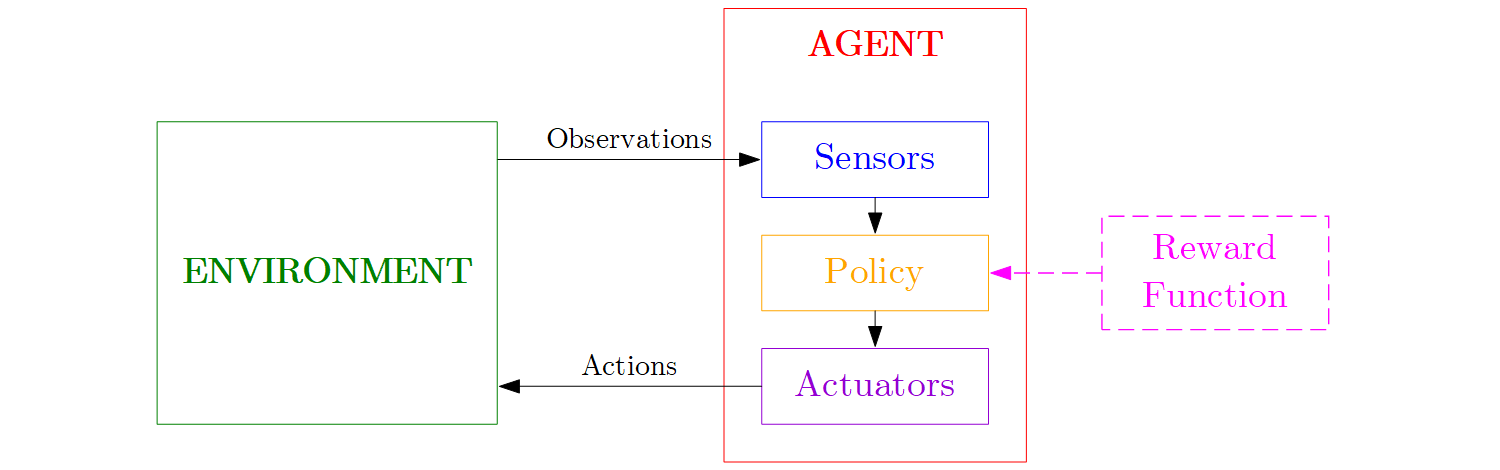}
	\caption{Reinforcement Learning Architecture}
	\label{fig:Reinforcement Learning Architecture}
	
\end{figure}

As depicted in figure \ref{fig:Reinforcement Learning Architecture}, an \textit{agent} trained using reinforcement learning technique essentially learns through trial and error. This allows the agent to explore the \textit{environment} and discover new strategies on its own, possibly some that may be even better than those of human beings!

Generally, the \textit{policy} $\pi$ to be trained for optimal behavior is a neural network, which acts as a function approximator governing the relationship between input \textit{observations} $o$ and output \textit{actions} $a$.

\begin{equation*}
\pi(a|o)
\end{equation*}

The agent is \textit{rewarded} for performing a set of ``good" actions and is \textit{penalized} for performing any ``bad" actions. The agent's goal is, therefore, to maximize the expected reward, given a specific policy.

\begin{equation*}
argmax\:E[R|\pi(a|o)]
\end{equation*}

Note that the term ``expected reward" is used instead of just ``reward". The expectation $E$, here, can be thought of as a simple average over a set of previous rewards based on a history of actions. While certain actions can produce exceedingly high/low reward points, it is a good habit to average them out in order generalize rewarding in the right direction. This reinforces the policy by updating its parameters such that ``generally good" actions are more likely to be performed in the future.

The most critical portion of reinforcement learning is, therefore, defining a \textit{reward function}. However, reward functions can be tricky to define. It is to be noted that the agent will perform any possible action(s) in order to maximize the reward, including cheating at times, essentially not learning the intended behavior at all. Defining a complex reward function, on the other hand, may limit the agent's creativity in terms of discovering novel strategies. Furthermore, rewarding the agent after performing good actions for a prolonged duration is also not a good idea, as the agent may wander randomly in the pursuit of maximizing its reward, but may likely never achieve the training objective.

That being said, let us see how the agent actually learns. At its core, there are two approaches to ``train" a policy, viz. working in the action space and working in the parameter space. The former approach converges faster while the later one provides a chance of better exploration.

\paragraph{\textbf{Working in Action Space}\\ \\}

\begin{figure}[htb]
	
	\centering
	\includegraphics[width=0.8\textwidth]{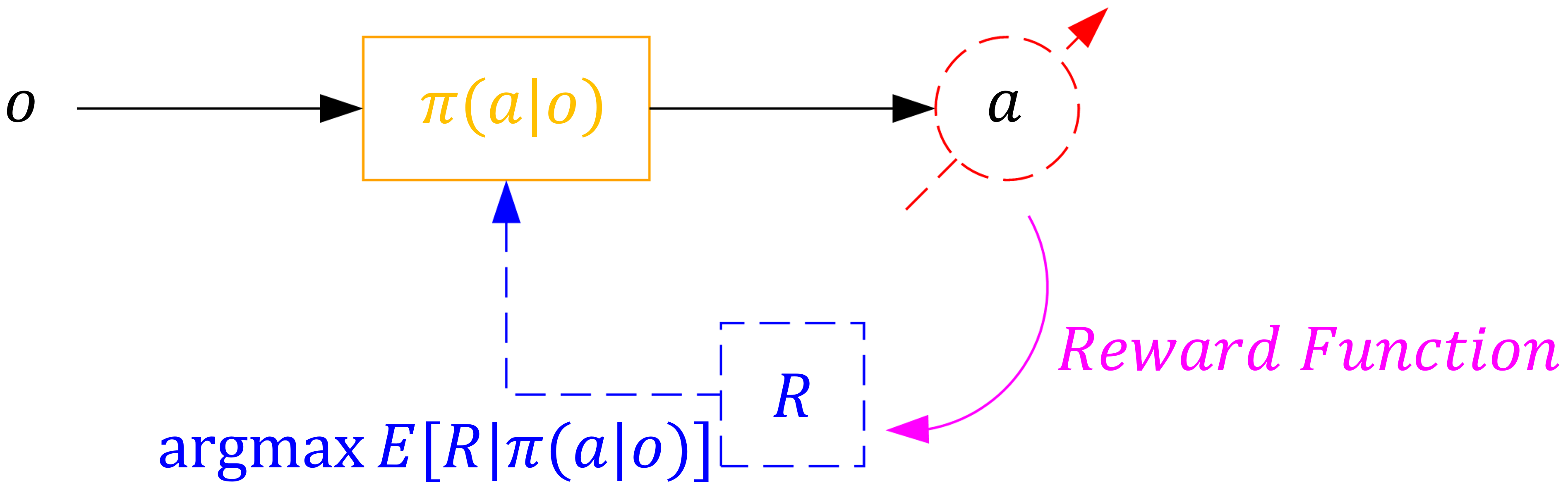}
	\caption{Reinforcement Learning in Action Space}
	\label{fig:Reinforcement Learning in Action Space}
	
\end{figure}

As depicted in figure \ref{fig:Reinforcement Learning in Action Space}, this approach deals with perturbing the action space and observing the reward being collected. The policy gradient can then be computed, which points towards the direction in which the policy parameters need to be updated.

One of the standard algorithms employing this approach is the \textit{proximal policy optimization (PPO)} algorithm. Initially, the agent performs random actions in order to explore the environment. The policy is recursively updated towards the rewarding actions using the \textit{gradient ascent} technique.

Instead of updating the policy regardless of how large its change was, PPO considers the probability ratio $r$ of old and new policies.

\begin{equation*}
r=\frac{\pi_{new}(a|o)}{\pi_{old}(a|o)}
\end{equation*}

In order to reduce variance, the reward function is replaced by a more general \textit{advantage function} $A$. The ``surrogate" objective function $L$ for PPO can be therefore written as follows.

\begin{equation*}
L=E\left[\frac{\pi_{new}(a|o)}{\pi_{old}(a|o)}*A\right]=E[r*A]
\end{equation*}

The PPO algorithm ensures that the updated policy $\pi_{new}$ may not be far away from the old policy $\pi_{old}$ by clamping the probability ratio $r$ within a suitable interval $[1-\epsilon,1+\epsilon]$, where $\epsilon$ is a hyperparameter, say, $\epsilon=0.2$. The main objective function $L_{clip}$ for PPO algorithm can be therefore written as follows.

\begin{equation*}
L_{clip}=E\left[min(r*A,clip(r,1-\epsilon,1+\epsilon)*A)\right]
\end{equation*}

Finally, the gradient ascent update rule for updating the policy parameters $\theta\:\in\:\mathbb{R}^d$ (weights and biases), considering a \textit{learning rate} hyperparameter $\alpha$ is defined as follows.

\begin{equation*}
\Delta\theta=\alpha*\nabla L_{clip}
\end{equation*}

While this approach increases stability and convergence rate, it reduces the agent's creativity and the agent may therefore get stuck in a sub-optimal solution.

\paragraph{\textbf{Working in Parameter Space}\\ \\}

\begin{figure}[htb]
	
	\centering
	\includegraphics[width=0.8\textwidth]{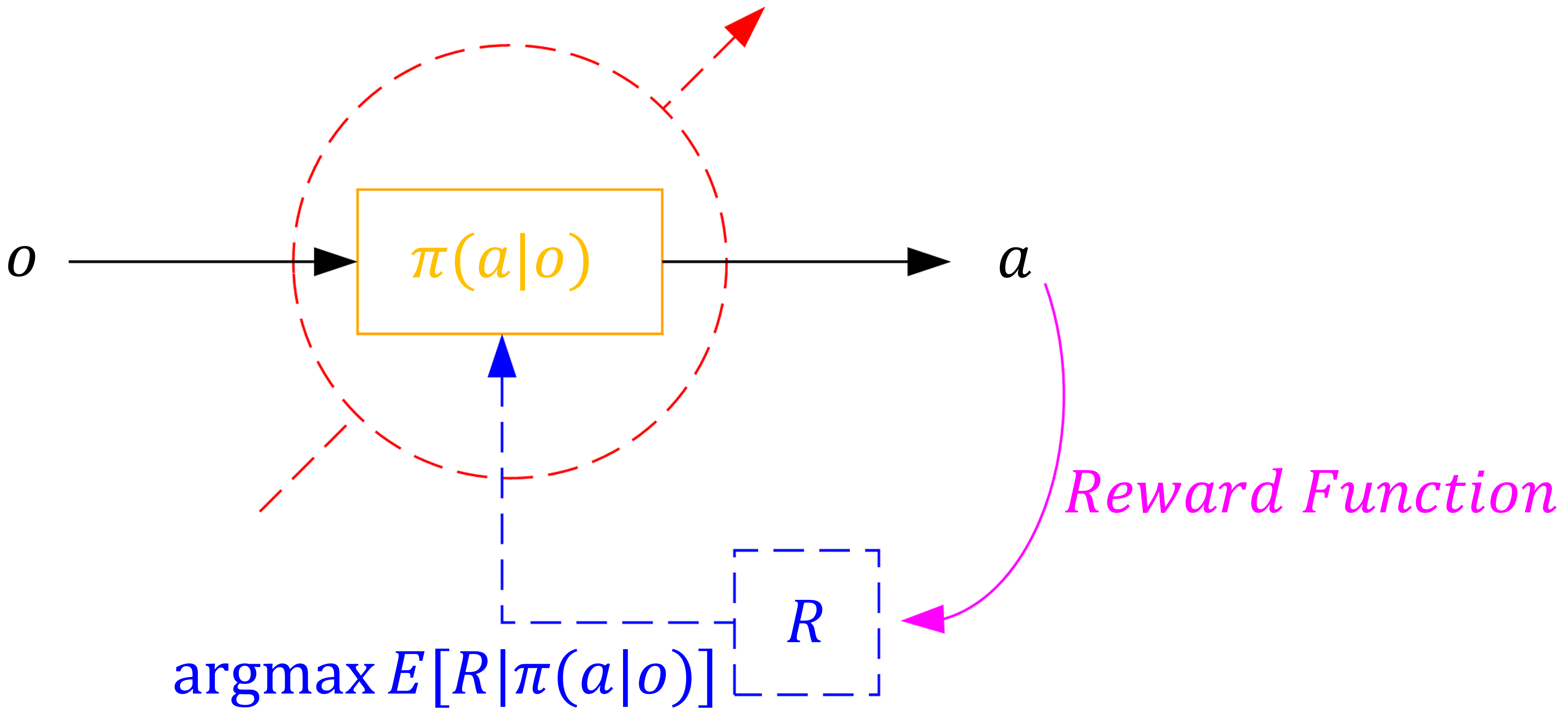}
	\caption{Reinforcement Learning in Parameter Space}
	\label{fig:Reinforcement Learning in Parameter Space}
	
\end{figure}

As depicted in figure \ref{fig:Reinforcement Learning in Parameter Space}, this approach deals with perturbing the parameter space directly and observing the reward being collected. This technique falls under the domains of \textit{genetic algorithms (GA)} or \textit{evolutionary strategies (ES)}.

Initially, a random population of policies is generated, which are all evaluated in order to analyze the reward obtained in each case. A new generation of policies is then populated in the direction with most promising results (maximum reward points), through the process of \textit{natural selection}. Random \textit{mutations} may also be introduced in order to take a chance towards creating a ``breakthrough" generation altogether. This process is repeated until an ``optimal" policy is obtained.

Since the policy is already updated as a part of exploration, this approach does not involve any gradient computation or back propagation, which makes it extremely easy to implement and ensures faster execution. Furthermore, this approach, almost every time, guarantees better exploration behavior as compared to the algorithms working in action space. Mathematically, this translates to a reduced risk of getting stuck in a local optima. However, algorithms working in the parameter space take a lot of time to converge. One has to, therefore, make a trade-off between convergence rate and exploration behavior, or design a hybrid approach addressing the problem at hand.

\pagebreak

\paragraph{\textbf{Reinforcement Learning Based Control Implementation}\\ \\}

\begin{figure}[htb]
	
	\centering
	\includegraphics[width=\textwidth]{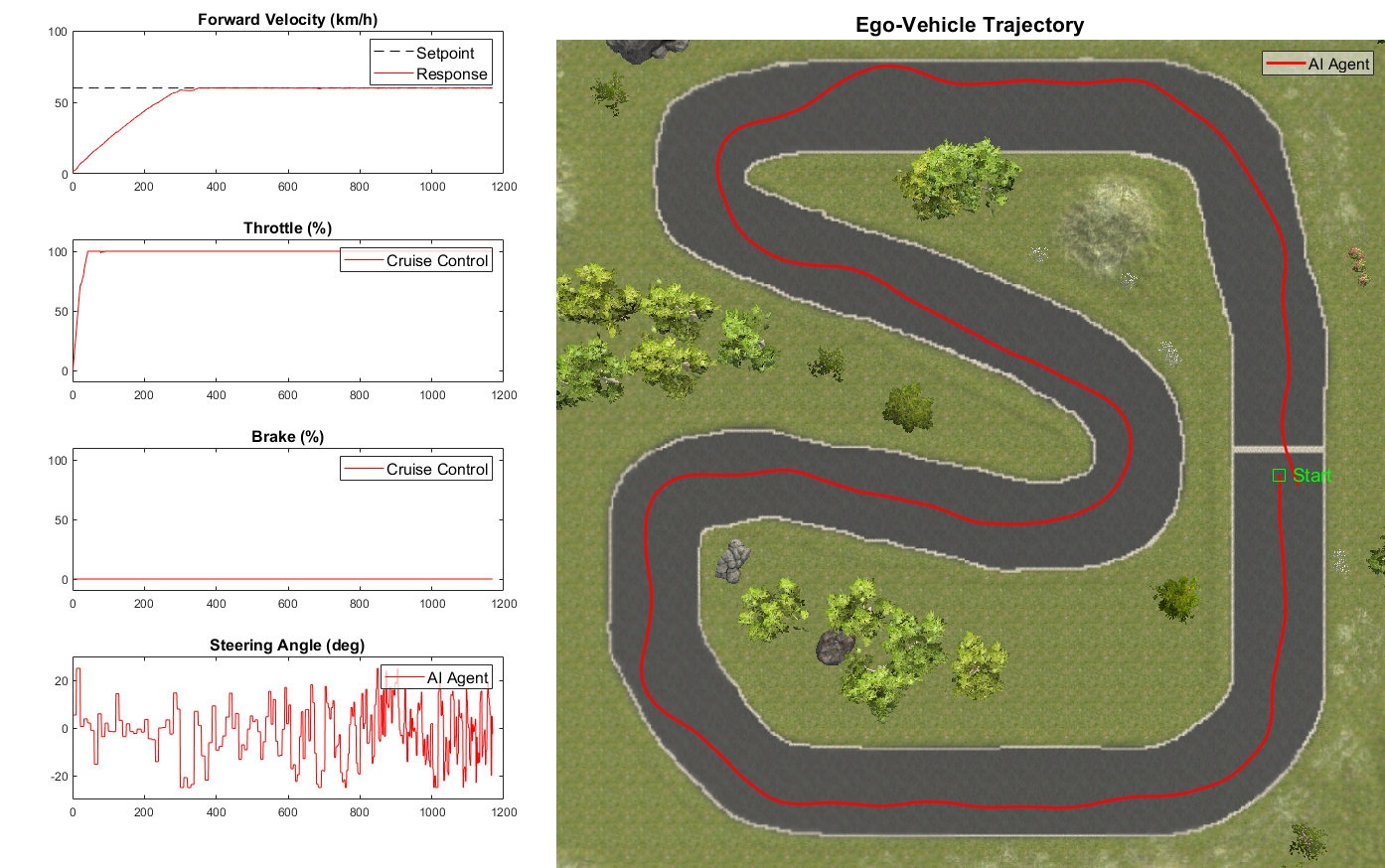}
	\caption{Reinforcement Learning Based Control Implementation}
	\label{fig:Reinforcement Learning Based Control Implementation}
	
\end{figure}

Implementation of reinforcement learning strategy for lateral control of a simulated autonomous vehicle is illustrated in figure \ref{fig:Reinforcement Learning Based Control Implementation}. Specifically, the longitudinal vehicle dynamics are controlled using a pre-tuned PID controller, while the lateral vehicle dynamics are controlled using a neural-network trained end-to-end (using the Proximal Policy Optimization algorithm).

Initially, the training was weighted towards completing an entire lap. The agent (ego vehicle) was rewarded proportional to its driving distance and was penalized for wandering off the road. The agent started off by performing random actions until it could get a sense of what was truly required in order to maximize its reward, and was finally able to complete an entire lap autonomously. The training was then weighted more towards reducing the lap time and the agent was rewarded inversely proportional to the lap time achieved and was additionally penalized in case the lap time was more than the average lap time achieved by human driver ($\sim48.3$ seconds). The final lap time achieved by AI agent trained using reinforcement learning was 47.2 seconds, which was better than the human driver!

Reinforcement learning has an upper hand when it comes to ``learning" unknown behaviors from scratch, through self-exploration. The policy adapts itself in order to maximize the expected reward, and explicit relations governing characteristics of the system are not necessary. Another peculiar advantage of this technique is its ability to discover novel strategies and even surpass human performance.

Nonetheless, owing to its exploratory nature, reinforcement learning requires a significantly high amount of training time, which might pose as a major issue in case of development-analysis cycles. Furthermore, this approach calls for a hand-crafted reward function, which, if defined inefficiently, may cause unintended rewarding, thereby leading to erroneous learning.

In conclusion, reinforcement learning is a powerful strategy, especially in terms of achieving super-human performance. However, it is generally better to have at least a basic pre-trained model available to work on top of, in order to skip the initial (and most time consuming) portion of the training phase, wherein the agent generates almost random actions with an aim of primary exploration.

\section{Our Research at Autonomous Systems Lab (SRMIST)}

Autonomous Systems Lab at the Department of Mechatronics Engineering, SRM Institute of Science and Technology (SRMIST) is headed by Mr. K. Sivanathan, and is focused on research and development in various aspects of autonomous systems. Specifically focusing on autonomous vehicles, our team is constantly working on all the three sub-systems viz. perception, planning and control. Although most of our past works are limited to simulation, we are currently working towards developing our own research platform for testing hardware implementations as well.

Further narrowing down to control of autonomous vehicles, our team has worked on most of the aforementioned control strategies, if not all, and we are continuously studying and experimenting with them in order to analyze their performance and test their limits. We are also working on developing novel and hybrid control strategies with an aim of ameliorating the controller performance (i.e. improved stability, tracking and robustness) and/or reducing computational cost (allowing real-time execution).

It is to be noted that an exhaustive discussion on each and every research project carried out at Autonomous Systems Lab (SRMIST) is beyond the scope of this chapter. Nonetheless, some of our team's research project demonstrations pertaining to autonomous vehicles are available at:

\noindent \url{www.youtube.com/playlist?list=PLdOCgvQ2Iny8g9OPmEJeVCas9tKr6auB3}

\section{Closing Remarks}

Throughout this chapter, we have discussed most of the current \textit{state-of-the-art} control strategies employed for autonomous vehicles, right from the basic ones such as bang-bang and PID control all the way up to advanced ones such as MPC and end-to-end learning, from theoretical, conceptual as well as practical viewpoints. Now, as we come towards the end of this chapter, we would like to add a few notes on some minute details pertaining to performance enhancement of control strategies for autonomous vehicles based on our knowledge and experience in the field.

\begin{itemize}
	
	\item \textbf{Gain Scheduling:} Most non-linear systems exhibit varying dynamics under different operating conditions. In such cases, a controller with single set of gains may not be able to regulate the system effectively throughout the operational design domain as the operating conditions change. The gain scheduling technique overcomes this limitation by employing multiple sets of gains appropriately across the entire range of operating conditions.
	
	\item \textbf{Integral Windup:} For PI and PID controllers, integrating the error over a prolonged duration may lead to very large values, resulting in loss of stability. It is, therefore, advised to limit the upper value of the accumulated error to a safe value. This can be achieved using a wide array of techniques, one of them being imposition of upper and lower bounds on the accumulated error and clipping the value within that range. Another approach would be limiting the size of history buffer over which the error is integrated through the use of a \textit{queue} data structure of limited size based on the first-in-first-out (FIFO) principle to discard error values that are too old.
	
	\item \textbf{Smoothing Controller Outputs:} Most of the controllers have a tendency of generating abrupt responses in case of a highly dynamic event, and while it may be appropriate in a few cases, it is generally undesirable. Such abrupt control actions may cause loss of stability or even affect the actuation elements adversely. It, therefore, becomes necessary to sooth the controller outputs before applying them to the plant, and one way of achieving this is to clamp the controller outputs within a safe range (determined through rigorous testing), such that the controller is still able to track the setpoint under all the operating conditions. A secondary approach would comprise of limiting the rate of change of the control output, thereby enabling a smoother transition.
	
	\item \textbf{Introduction of Dead-Bands:} Most corrective controllers oscillate about the setpoint, trying really hard to converge, and while this might be desirable in case of highly precise systems, it is generally not really necessary to achieve the setpoint exactly. The controller action may be nullified within close neighborhood of the setpoint. This is known as a dead-band. It must be noted, however, that the width of the dead-band must be chosen wisely, depending upon the precision requirements of the system in question.
	
	\item \textbf{Safe Reinforcement Learning:} As discussed earlier, reinforcement learning involves discovery of novel strategies majorly through a trial-and-error approach, due to which, it is seldom directly applied to a physical system (you don't want to take millions of episodes to learn by crashing the agent almost each time). However, there are a few techniques that can be employed to overcome this limitation, one of them being training a simulated agent under pseudo-realistic operating conditions and then transferring the trained model onto the physical agent. Another way of approaching this problem would be to penalize the policy prior to any catastrophic event by self-supervising the agent through an array of sensors. A third (and more elegant) approach would be to train the agent for a ``safe" behavior using imitation learning and then improving its performance by training it further using reinforcement learning.

\end{itemize}

\begin{table}[h]
	\caption{Summary of Control Strategies for Autonomous Vehicles}
	\label{Summary of Control Strategies for Autonomous Vehicles}
	\centering
	\resizebox{\textwidth}{!}{%
		\begin{tabular}{|l|c|c|c|c|c|c|}
			\hline
			\multicolumn{1}{|c|}{\textbf{Comparison Metric}} & \textbf{\begin{tabular}[c]{@{}c@{}}Bang-Bang\\ Control\end{tabular}} & \textbf{\begin{tabular}[c]{@{}c@{}}PID\\ Control\end{tabular}}      & \textbf{\begin{tabular}[c]{@{}c@{}}Geometric\\ Control\end{tabular}} & \textbf{\begin{tabular}[c]{@{}c@{}}Model Predictive\\ Control\end{tabular}} & \textbf{\begin{tabular}[c]{@{}c@{}}Imitation Learning\\ Based Control\end{tabular}} & \textbf{\begin{tabular}[c]{@{}c@{}}Reinforcement Learning\\ Based Control\end{tabular}} \\ \hline
			Principle of Operation                           & \begin{tabular}[c]{@{}c@{}}Error Driven\\ (FSM)\end{tabular}         & \begin{tabular}[c]{@{}c@{}}Error Driven\\ (Correcitve)\end{tabular} & \begin{tabular}[c]{@{}c@{}}Model Based\\ (Corrective)\end{tabular}   & \begin{tabular}[c]{@{}c@{}}Model Based\\ (Optimal)\end{tabular}             & \begin{tabular}[c]{@{}c@{}}Supervised\\ Learning\end{tabular}                       & \begin{tabular}[c]{@{}c@{}}Reinforcement\\ Learning\end{tabular}                        \\ \hline
			Tracking                                         & Poor                                                                 & Good                                                                & Very Good                                                            & Excellent                                                                   & Good                                                                                & -                                                                                       \\ \hline
			Robustness                                       & Poor                                                                 & Very Good                                                           & Very Good                                                            & Excellent                                                                   & Good                                                                                & Good                                                                                    \\ \hline
			Stability                                        & Poor                                                                 & Very Good                                                           & Very Good                                                            & Excellent                                                                   & Good                                                                                & Good                                                                                    \\ \hline
			Reliability                                      & Very Low                                                             & Very High                                                           & Very High                                                            & Extremely High                                                              & Low                                                                                 & Low                                                                                     \\ \hline
			Technical Complexity                             & Very Low                                                             & Low                                                                 & Low                                                                  & Very High                                                                   & Low                                                                                 & High                                                                                    \\ \hline
			Computational Overhead                           & Very Low                                                             & Low                                                                 & Low                                                                  & Very High                                                                   & High                                                                                & High                                                                                    \\ \hline
			Constraint Satisfaction                          & Poor                                                                 & Good                                                                & Good                                                                 & Excellent                                                                   & -                                                                                   & -                                                                                       \\ \hline
			Technical Maturity                               & Very High                                                            & Very High                                                           & High                                                                 & Moderate                                                                    & Low                                                                                 & Very Low                                                                                \\ \hline
		\end{tabular}%
	}
\end{table}

As a final note, we summarize the control strategies described in this chapter against a set of qualitative comparison metrics in table \ref{Summary of Control Strategies for Autonomous Vehicles}.

Generally, when it comes to traditional control strategies, PI or PID controllers are the primary choice for longitudinal control, whereas MPC or geometric controllers are predominantly employed for lateral control. However, exact choice depends upon the specifics of the problem statement. Learning based control strategies, on the other hand, do not have any ``preferred" implementations, and it would be appropriate to say that these techniques are currently in their infant stage with a bright future ahead of them.

\begin{thefurtherreading}{99}

% Reference Book on Control Engineering
\bibitem{} Ogata, K. (2010). \textit{Modern Control Engineering.} \nth{5} Edition, Prentice Hall.

% Modelling
\bibitem{} Kong, J., Pfeiffer, M., Schildbach, G., and Borrelli, F. (2015). Kinematic and Dynamic Vehicle Models for Autonomous Driving Control Design. \textit{IEEE Intelligent Vehicles Symposium (IV)}, Seoul, pp. 1094--1099.

% Modelling
\bibitem{} Polack, P., Altche, F., Novel, B.A., and de La Fortelle, A. (2017). The Kinematic Bicycle Model: A Consistent Model for Planning Feasible Trajectories for Autonomous Vehicles? \textit{IEEE Intelligent Vehicles Symposium (IV)}, Redondo Beach, pp. 812--818.

% Bang-Bang Control
\bibitem{} O'Brien, R.T. (2006). Bang-Bang Control for Type-2 Systems. \textit{Proceeding of the Thirty-Eighth Southeastern Symposium on System Theory}, Cookeville, pp. 163--166.

% PID Control
\bibitem{} Emirler, M.T., Uygan, I.M.C., Guvenc, B.A., and Guvenc, L. (2014). Robust PID Steering Control in Parameter Space for Highly Automated Driving. \textit{International Journal of Vehicular Technology}, vol. 2014.

% Pure Pursuit Control
\bibitem{} Coulter, R.C. (1992). Implementation of the Pure Pursuit Path Tracking Algorithm. \textit{CMU-RI-TR-92-01}.

% Stanley Control
\bibitem{} Hoffmann, G.M., Tomlin, C.J., Montemerlo, M., and Thrun, S. (2007). Autonomous Automobile Trajectory Tracking for Off-Road Driving: Controller Design, Experimental Validation and Racing. \textit{American Control Conference}, New York, pp. 2296--2301.

% Autonomous Steering Controllers
\bibitem{} Snider, J.M. (2009). Automatic Steering Methods for Autonomous Automobile Path Tracking. \textit{CMU-RI-TR-09-08}.

% Model Predictive Control
\bibitem{} Zanon, M., Frasch, J., Vukov, M., Sager, S., and Diehl, M. (2014). \textit{Model Predictive Control of Autonomous Vehicles}. In: Waschl H., Kolmanovsky I., Steinbuch M., del Re L. (eds) Optimization and Optimal Control in Automotive Systems. Lecture Notes in Control and Information Sciences, vol. 455, Springer, Cham.

% Model Predictive Control
\bibitem{} Babu, M., Theerthala, R.R., Singh, A.K., Baladhurgesh B.P., Gopalakrishnan, B., and Krishna, K.M. (2019). Model Predictive Control for Autonomous Driving considering Actuator Dynamics. \textit{American Control Conference (ACC)}, Philadelphia, pp. 1983--1989.

% Model Predictive Control
\bibitem{} Obayashi, M., and Takano, G. (2018). Real-Time Autonomous Car Motion Planning using NMPC with Approximated Problem Considering Traffic Environment. \textit{6th IFAC Conference on Nonlinear Model Predictive Control NMPC}, IFAC-PapersOnLine, vol. 51, no. 20, pp. 279--286.

% End-To-End Learning Based Control
\bibitem{} Tampuu, A., Semikin, M., Muhammad, N., Fishman, D., and Matiisen, T. (2020). A Survey of End-to-End Driving: Architectures and Training Methods. \textit{Preprint}, arXiv:2003.06404.

% Imitation Learning Based Control
\bibitem{} Pomerleau, D. (1989). ALVINN: An Autonomous Land Vehicle In a Neural Network. \textit{Proceedings of Advances in Neural Information Processing Systems 1}, pp. 305--313.

% Imitation Learning Based Control
\bibitem{} LeCun, Y., Muller, U., Ben, J., Cosatto, E., and Flepp, B. (2005). Off-Road Obstacle Avoidance through End-to-End Learning. \textit{Advances in Neural Information Processing Systems}, pp. 739--746.

% Imitation Learning Based Control
\bibitem{} Bojarski, M., Testa, D.D., Dworakowski, D., Firner, B., Flepp, B., Goyal, P., Jackel, L.D., Monfort, M., Muller, U., Zhang, J., Zhang, X., Zhao, J., and Zieba, K. (2016). End to End Learning for Self-Driving Cars. \textit{Preprint}, arXiv:1604.07316.

% Imitation Learning Based Control
\bibitem{} Samak, T.V., Samak, C.V., and Kandhasamy, S. (2020). Robust Behavioral Cloning for Autonomous Vehicles using End-to-End Imitation Learning. \textit{Preprint}, arXiv:2010.04767.

% Reinforcement Learning Based Control
\bibitem{} Li, D., Zhao, D., Zhang, Q., and Chen, Y. (2019). Reinforcement Learning and Deep Learning Based Lateral Control for Autonomous Driving [Application Notes]. \textit{IEEE Computational Intelligence Magazine}, vol. 14, no. 2, pp. 83--98.

% Reinforcement Learning Based Control
\bibitem{} Riedmiller, M., Montemerlo, M., and Dahlkamp, H. (2007). Learning to Drive a Real Car in 20 Minutes. \textit{Frontiers in the Convergence of Bioscience and Information Technologies}, Jeju City, pp. 645--650.

% Reinforcement Learning Based Control
\bibitem{} Kendall, A., Hawke, J., Janz, D., Mazur, P., Reda, D., Allen, J.M., Lam, V.D., Bewley, A., and Shah, A. (2018). Learning to Drive in a Day. \textit{Preprint}, arXiv:1807.00412.

% Learning Based MPC
\bibitem{} Kabzan, J., Hewing, L., Liniger, A., and Zeilinger, M.N. (2019). Learning-Based Model Predictive Control for Autonomous Racing. \textit{IEEE Robotics and Automation Letters}, vol. 4, no. 4, pp. 3363--3370.

% Learning Based MPC
\bibitem{} Rosolia, U., and Borrelli, F. (2020). Learning How to Autonomously Race a Car: A Predictive Control Approach. \textit{IEEE Transactions on Control Systems Technology}, vol. 28, no. 6, pp. 2713--2719.

% Hybrid Control
\bibitem{} Drews, P., Williams, G., Goldfain, B., Theodorou, E.A., and Rehg, J.M. (2018). Vision-Based High Speed Driving with a Deep Dynamic Observer. \textit{Preprint}, arXiv:1812.02071.

\end{thefurtherreading}

\end{document}